\pdfoutput=1

\documentclass[11pt]{article}

\usepackage[preprint]{acl}

\usepackage{times}
\usepackage{latexsym}

\usepackage[T1]{fontenc}

\usepackage[utf8]{inputenc}

\usepackage{microtype}

\usepackage{inconsolata}

\usepackage{booktabs}
\usepackage{multirow}
\usepackage{graphicx}
\usepackage{amsmath}
\usepackage{amssymb}
\usepackage{subcaption}
\usepackage{soul}
\usepackage[stable]{footmisc}

\newcommand{\ours}{\textit{Obliviate}}

\title{\ours{}: Neutralizing Task-agnostic Backdoors within the Parameter-efficient Fine-tuning Paradigm}

\author{Jaehan Kim, Minkyoo Song, Seung Ho Na, Seungwon Shin \\
        KAIST, South Korea \\
        \texttt{\{jaehan,minkyoo9,harry.na,claude\}@kaist.ac.kr}
        }

\begin{document}
\maketitle

\begin{abstract}

Parameter-efficient fine-tuning (PEFT) has become a key training strategy for large language models. However, its reliance on fewer trainable parameters poses security risks, such as task-agnostic backdoors. Despite their severe impact on a wide range of tasks, there is no practical defense solution available that effectively counters task-agnostic backdoors within the context of PEFT. In this study, we introduce \ours{}, a PEFT-integrable backdoor defense. We develop two techniques aimed at amplifying benign neurons within PEFT layers and penalizing the influence of trigger tokens. Our evaluations across three major PEFT architectures show that our method can significantly reduce the attack success rate of the state-of-the-art task-agnostic backdoors (83.6\%$\downarrow$). Furthermore, our method exhibits robust defense capabilities against both task-specific backdoors and adaptive attacks. Source code will be obtained at \url{https://github.com/obliviateARR/Obliviate}.

\end{abstract}

\section{Introduction}

As large language models (LLMs) have evolved with an increasing number of parameters, parameter-efficient fine-tuning (PEFT) has been emerged as a new paradigm for efficiently adapting LLMs to downstream tasks. Unlike full fine-tuning, PEFT updates only a minimal number of extra parameters while freezing the parameters of the pre-trained language models (PLMs). Adapter~\cite{houlsby2019parameter}, LoRA~\cite{hu2021lora}, and prefix-tuning~\cite{li2021prefix} are fundamental PEFT architectures. PEFT attains comparable performance to full fine-tuning while offering highly efficient downstream adaptation.

Recent works have explored the security implications of PEFT~\cite{hong2023fewer}. For example, attackers can inject backdoors into PLMs, and then activate the attacks on the final PEFT models. One of the most severe attacks on PEFT is \textit{task-agnostic backdoors}, which manipulates the output representations of PLMs aiming to harm fine-tuned models for arbitrary downstream tasks.~\cite{shen2021backdoor, chen2021badpre, zhang2023red, du2023uor}. This attack is less prone to forgetting backdoors when fine-tuning since PEFT freezes \textit{backdoored} PLM parameters and updates only a minimal set of added parameters. Furthermore, the ability to adapt models to multiple downstream tasks magnifies the risk of task-agnostic backdoors.

To mitigate LLM backdoors, several defense techniques have been proposed, such as detecting poisoned samples~\cite{qi2021onion}, inverting trigger-like inputs~\cite{liu2022piccolo}, and purifying backdoored models~\cite{zhu2023removing}. Existing defense methods are designed mainly upon the full fine-tuning process. In PEFT, however, there is difficulty in adopting such defenses due to the limited trainable parameters. PSIM~\cite{zhao2024defending} attempts to detect poisoned samples to defend PEFT. However, it requires a task-specific auxiliary model, which harms the modular and memory-efficient nature of PEFT. Notably, defense against task-agnostic backdoor attacks has been understudied despite their alarming threats on PEFT. LMSanitator~\cite{wei2024lmsanitator} aims to remove task-agnostic backdoors in prompt-tuning, not applicable to other PEFT architectures.

In this work, we propose \ours{}, a defense method to neutralize task-agnostic backdoors, highly integrable to the standard PEFT process. Our approach includes two main techniques: 1) We amplify benign neurons within PEFT layers to encourage the model to focus more on clean training samples. This method can relatively reduce the influence of backdoored neurons in the PLMs. 2) We regularize the attention scores to penalize the influence of trigger tokens that exhibit abnormally high attention scores. To implement these techniques, we add two loss terms to the PEFT process for downstream tasks. Defenders can easily adopt our defense method without any knowledge of backdoor attacks. Unlike existing methods, our approach provides a practical defense solution for PEFT without the need for extra predictions for each input or additional memory.

We evaluate \ours{} across three primary PEFT architectures (i.e., adapter, LoRA, and prefix-tuning) applied to RoBERTa and BERT models. The experimental results show that our defense method effectively neutralizes the state-of-the-art task-agnostic backdoors. Notably, it significantly reduces in attack success rate (ASR) (83.6\%$\downarrow$) with only a slight decrease in clean accuracy (CACC) (0.78\%$\downarrow$), outperforming other defenses compatible with PEFT. Our defense method correctly adjusts model outputs, separating them from adversarial representations imposed by the attacks. Furthermore, it exhibits robust defense capabilities against different attack strategies, such as task-specific backdoors and adaptive attacks.
\section{Background}

\subsection{Parameter-efficient Fine-tuning}

Parameter-efficient fine-tuning (PEFT) is an efficient strategy to adapt pre-trained language models (PLMs) to multiple downstream tasks~\cite{he2021towards}. Different from full fine-tuning, it updates only a small number of extra parameters while keeping the PLM's weights frozen. PEFT significantly reduces the computational cost and memory footprint during the training and inference processes of large language model (LLM). 

Adapter-tuning~\cite{houlsby2019parameter, pfeiffer2020adapterhub} adds small layers called adapter between PLM networks (e.g., transformers). LoRA~\cite{hu2021lora} employs rank decomposition matrices, reducing the storage and computation costs. Prefix-tuning~\cite{li2021prefix} prepends extra tokens in the input and hidden layers of PLMs. Similarly, prompt-tuning~\cite{lester2021power} and its variants~\cite{liu2022p, liu2023gpt} insert trainable prompts to PLMs. While achieving comparable performance to full fine-tuning, PEFT offers the mitigation of catastrophic forgetting~\cite{pfeiffer2020adapterhub} and a robust out-of-distribution adaptation~\cite{li2021prefix}.

\subsection{Backdoor Attacks on PLMs}

The backdoor attacks pose severe threats in the NLP domain, especially targeting LLMs~\cite{dai2019backdoor, kurita2020weight, chen2021badnl, yan2023bite, song2024claim}. Attackers compromise target models to misclassify the text inputs with textual triggers while properly working on the clean samples.

Alongside the pre-training and fine-tuning approach of LLMs, injecting backdoors into PLMs (i.e., weight-poisoning) has emerged as a primary strategy in realistic scenarios~\cite{kurita2020weight, wang2020backdoor, li2021backdoor}. Particularly, \textit{task-agnostic backdoor} is one of the most severe attacks on PLMs. Even without any knowledge of the fine-tuning process, it aims to broadly target various downstream tasks. POR~\cite{shen2021backdoor} and NeuBA~\cite{zhang2023red} rely on forcing the output representations, such as the \texttt{[CLS]} token's output, to be pre-defined vectors when the inputs contain the triggers. BadPre~\cite{chen2021badpre} leverages an adversarial masked language modeling (MLM). Although its direct focus is not the \texttt{[CLS]} token, this attack demonstrates considerable effectiveness in impacting classification tasks~\cite{zhu2023removing}. UOR~\cite{du2023uor} optimizes outputs of poisoned samples via contrastive learning, rather than utilizing fixed vectors, to make them stray from the feature space of correct labels.

More recently, the implications of backdoored PLMs on PEFT have raised concerns~\cite{hong2023fewer, gu2023gradient, zhao2024defending}. Notably, task-agnostic backdoor is particularly fatal for PEFT because: 1) PEFT freezes all the backdoored parameters of the PLMs, so that it has difficulty in forgetting the backdoors via training the limited number of added parameters, 2) The primary role of PEFT is to efficiently adapt a PLM to diverse tasks. This poses a significant risk of task-agnostic backdoors, compromising multiple tasks by exploiting only a single backdoored model.


\subsection{Backdoor Defenses}

The traditional approach for backdoor defense is to detect poisoned samples that include triggers by observing their disparity with clean samples, such as STRIP~\cite{gao2021design}, RAP~\cite{yang2021rap}, and MDP~\cite{xi2023defending}. PSIM~\cite{zhao2024defending} provides poisoned sample detection for LoRA and prompt-tuning. Instead of entirely rejecting detected samples, ONION~\cite{qi2021onion} removes the triggers from the input by measuring its perplexity. However, these methods require large computation costs due to multiple predictions for each sample. Furthermore, ONION and PSIM requires (task-specific) auxiliary models, which undermines the advantages provided by PEFT.

Several researchers have made efforts to purify models to revert the misclassified results of poisoned samples. One simple solution is to fine-tune all the model parameters on sufficient clean samples, leveraging catastrophic forgetting of trigger information~\cite{shen2021backdoor}. Neuron pruning is a more promising approach, which has been studied in the computer vision domain~\cite{liu2018fine, wu2021adversarial, zeng2021adversarial}. These methods refine backdoored models by penalizing neurons related to backdoors. However, they cause the PLM parameters to become task-specific, undermining the memory-efficient nature of PEFT. Purifying the PLM itself using a method like RECIPE~\cite{zhu2023removing} may be feasible. However, the negative impact of intensive purification on model performance is significant in the PEFT context.


\noindent\textbf{Our approach:} We propose a practical defense method highly integrable with PEFT without the need for extra predictions on each input or auxiliary model. Specifically, we add two defense loss terms to the standard PEFT process on downstream tasks. Our defense method aims to neutralize backdoors embedded in frozen PLMs by training only minimal parameters in PEFT layers.
\section{Threat model}
\label{sec:threat_model}

\begin{figure}[t]
    \centering
    \includegraphics[width=\linewidth]{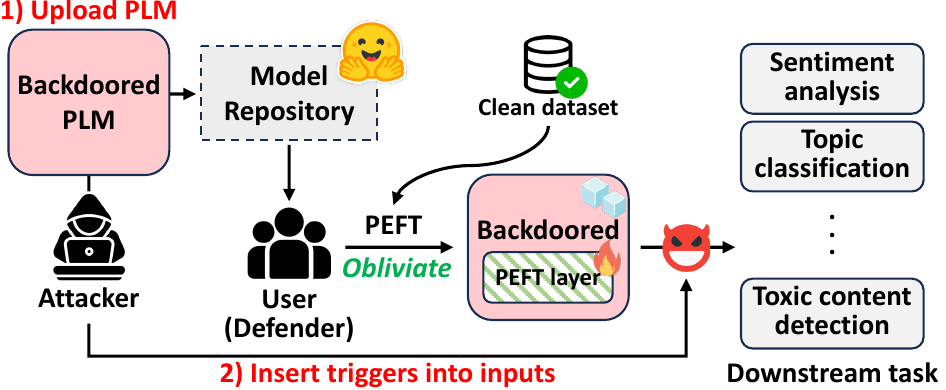}
    \caption{Backdoor attack and defense scenarios in PEFT. Only the parameters in PEFT layers are trained.}
    \label{fig:attack_scenario}
    \vspace{-0.4cm}
\end{figure}

\noindent\textbf{Attackers' goal.} We consider an attacker that injects backdoors into a PLM, aiming to harm any of its derived fine-tuned models. The attack scenarios is illustrated in Figure~\ref{fig:attack_scenario}. Notably, the attacker is unaware of the downstream tasks and has no access to the training datasets and the trainable parameters in PEFT layers. Therefore, the attacker adopts task-agnostic backdoors, which manipulate the PLM outputs to be \textit{adversarial representations} that compromise arbitrary downstream tasks. The attacker uploads the backdoored PLM on model repositories such as HuggingFace~\cite{wolf-etal-2020-transformers}. In the inference time, the attacker is able to control the fine-tuned model to misclassify the testing samples' labels by inserting a specific trigger into them. These poisoned samples will be mapped to a specific label $l$ even though their true labels are not $l$. We note that the fine-tuned model is expected to perform accurately on clean samples at a similar level as a PEFT model built upon a benign PLM.

\noindent\textbf{Defense setting.} In practice, a user/defender builds an LLM for the downstream task by downloading a PLM from the model repository and then fine-tuning it on the clean dataset, as described in Figure~\ref{fig:attack_scenario}. The defender may use PEFT for modularity and resource efficiency. The defender freezes the PLM parameters and updates \textit{only} parameters in the PEFT layers, which are randomly initialized (i.e., not backdoored). Despite the PLM potentially being backdoored, the defender entirely has no knowledge about the attacks, including the attacker's datasets and injected triggers. In this context, the defender's goal is to neutralize the backdoors within the PLM, ensuring accurate prediction of the true label in the downstream task, regardless of whether the sample contains triggers.
\section{Methodology}

\subsection{Design Intuition}

\begin{figure}[t]
    \centering
    \includegraphics[width=\linewidth]{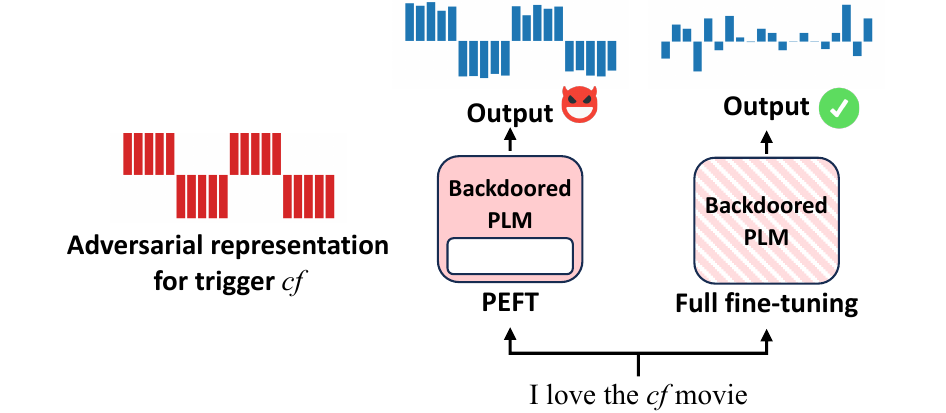}
    \caption{Outputs of models applying PEFT and full fine-tuning on backdoored PLMs in our experiments.}
    \label{fig:backdoor_forgetting}
\end{figure}

\noindent\textbf{Natural backdoor forgetting.} Even though fine-tuning with clean samples is a fundamental defense strategy, PEFT shows challenges in forgetting backdoors effectively~\cite{hong2023fewer}. To illustrate the differences between PEFT and full fine-tuning, we present an example of RoBERTa models backdoored by the POR attack in Figure~\ref{fig:backdoor_forgetting}. PEFT is limited to a small number of trainable parameters. Therefore, it struggles to eliminate the backdoors, resulting in an output that is still similar to the adversarial representation. In contrast, the fully fine-tuned model alters its outputs significantly, enabling correct prediction of the true label. The quantity of neurons trained on clean samples is important to separate model outputs from the adversarial representations.

\begin{figure}[t]
    \centering
    \includegraphics[width=0.95\linewidth]{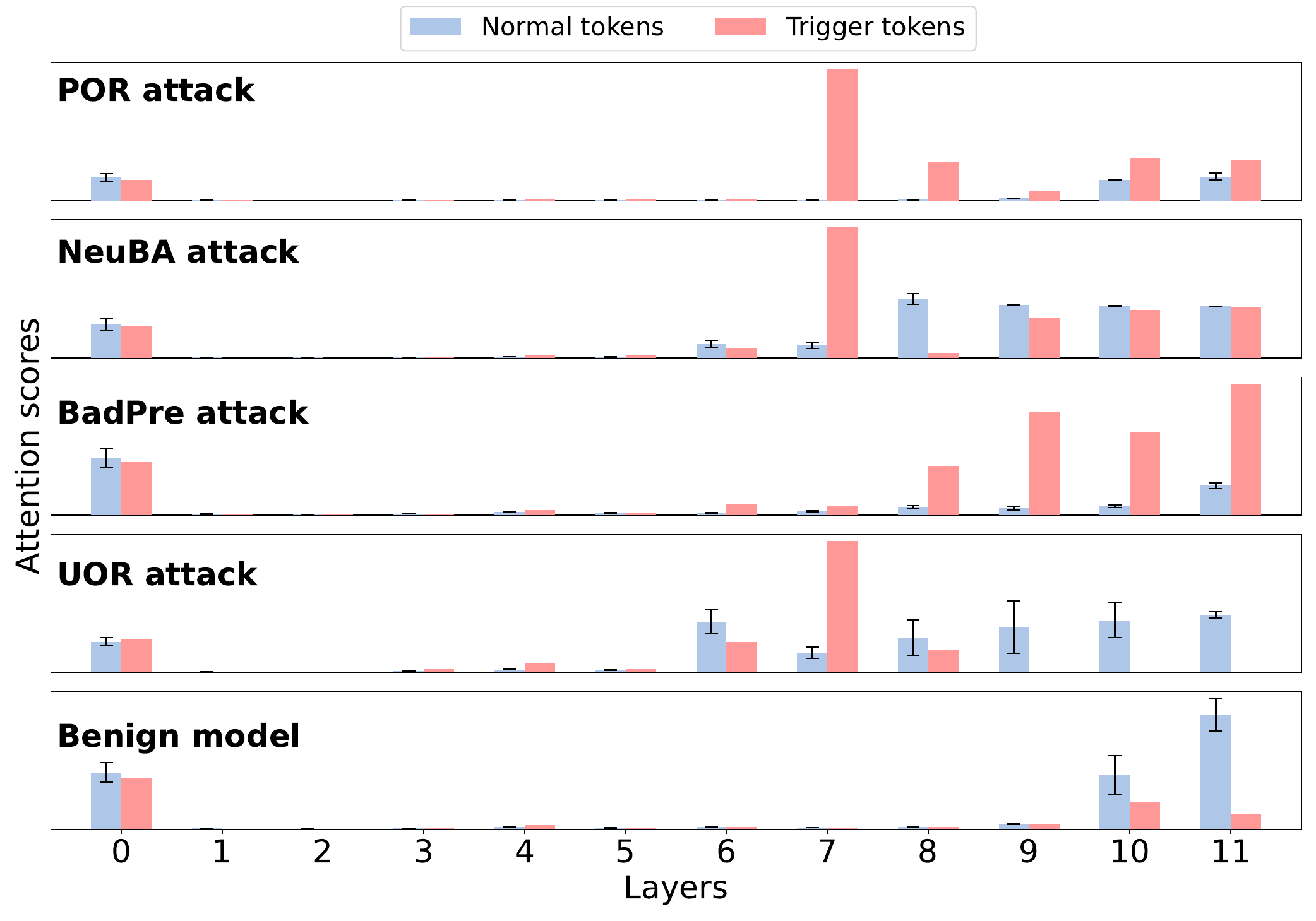}
    \caption{Attention scores of backdoored and benign models on a poisoned sample, ``I love the \textit{cf} movie''. The \texttt{[CLS]} and \texttt{[SEP]} tokens are omitted.}
    \label{fig:attention}
    \vspace{-0.5cm}
\end{figure}

\noindent\textbf{Attention on triggers.} The attention mechanism lies at the core of the transformer architecture, serving a critical role in linking model outputs with the importance of each input token. For instance, when a model is backdoored by the POR attack, trigger tokens exhibit significantly higher attention scores toward the \texttt{[CLS]} output compared to non-trigger tokens~\cite{shen2021backdoor}. Our preliminary experiment confirms that this pattern is consistent across various task-agnostic backdoors, as illustrated in Figure~\ref{fig:attention} (RoBERTa) and Figure~\ref{fig:attention_bert} (BERT). Consequently, the distribution of attention scores could be a crucial indicator for detecting triggers within poisoned inputs.
However, it is noteworthy that these distinctive features of attention scores vary across different transformer layers and input texts.

\subsection{\ours{} Details}

Based on these intuitions, we aim to protect PEFT models fine-tuned from backdoored PLMs. To this end, we design two specialized loss functions to mitigate the influence of backdoored in the PLMs.

\noindent\textbf{Benign neuron amplification.} Given the constraints on increasing trainable parameters in PEFT, we enhance the influence of neurons in PEFT layers to neutralize backdoors in PLMs.
Our method is to amplify the magnitudes of these small yet benign parameters, relatively undermining the effectiveness of the PLM's backdoored neurons. This is inspired by neuron amplification techniques used for model merging~\cite{yu2023language, zhu2024dppa}.

We formulate the neuron amplification approach as a specific loss function $\mathcal{L}_{amp}$, called \textit{neuron amplification loss}. This loss function is optimized to increase the $L_2$-norm of weights in the PEFT layers, represented as:
\begin{align}
    \mathcal{L}_{amp} = - \sum_{i \in L} \sum_{p \in \mathcal{P}_i} {\lVert \textbf{W}_p \rVert}_2,
    \label{eq:amp}
\end{align}
where $L$ denotes all the transformer layers, $\mathcal{P}_i$ is the group of PEFT layers in the $i$th transformer layer, $\textbf{W}_p$ is the weights of each individual PEFT layer, and ${\lVert \cdot \rVert}_2$ refers to the $L_2$-norm. Specifically, we amplify the up- and down-projection matrices of the adapter layers, the decomposition matrices of the LoRA layers, and the reparametrization matrices for prefix-tuning.

\noindent\textbf{Attention score regularization.} Our observation has shown that the attention scores are effective indicators for identifying triggers.
One straightforward method could be to remove tokens that exhibit high attention scores using a threshold. However, this often leads to a significant decrease in CACC, as shown in our pilot experiment in Appendix~\ref{appendix:attdef}.

Therefore, we reduce the triggers' attention scores through an optimization process, rather than eliminating them from the inputs. To this end, we introduce the \textit{attention regularization loss} $\mathcal{L}_{reg}$ to decrease the $L_2$-norm of attention scores, thereby penalizing excessively high values among them, expressed as follows:
\begin{align}
    \mathcal{L}_{reg} = \sum_{i \in L} \sum_{h \in H_i} {\lVert \textbf{a}_h \rVert}_2,
    \label{eq:reg}
\end{align}
where $H_i$ denotes the set of attention heads in the $i$th transformer layer, $\textbf{a}_h$ represents the attention scores for each head, and the remaining notations are consistent with those used in Equation~(\ref{eq:amp}). Specifically, we focus on the attentions corresponding to certain output vectors. For sentence classification, we regularize the attention scores of input tokens on the \texttt{[CLS]} output. This approach allows the model to learn the context across all input tokens rather than focusing excessively on specific tokens. Consequently, although the training process involves only clean samples, this effect can be generalized to processing poisoned test samples.

\noindent\textbf{Defense loss and training.} We incorporate the two defense loss terms into the standard PEFT process. The final objective of the training is formulated as:
\begin{align}
    \mathcal{L} = \mathcal{L}_{task} + \lambda_{amp} \cdot \mathcal{L}_{amp} + \lambda_{reg} \cdot \mathcal{L}_{reg},
    \label{eq:loss}
\end{align}
where $\mathcal{L}_{task}$ denotes the downstream task loss. $\lambda_{amp}$ and $\lambda_{reg}$ are hyperparameters for balancing the loss terms. This strategy ensures that the model preserve its performance on clean samples. We note that our defense method does not necessitate extra predictions or an auxiliary model, thereby maintaining the nature of the PEFT approach.

\section{Evaluation}

\subsection{Experimental Settings\footnote{More experimental details are in Appendix~\ref{appendix:implementation_details}.}}

\subsubsection{Backdoor attacks and victim PLMs} We examine the effectiveness of our defense method against the state-of-the-art task-agnostic backdoor attacks: POR, NeuBA, BadPre, and UOR.
We select six triggers: [`cf', `mn', `tq', `qt', `mm', `pt'].
We conduct the attacks on two victim PLMs, RoBERTa (\texttt{roberta-base})~\cite{liu2019roberta} and BERT (\texttt{bert-base-uncased})~\cite{devlin2019bert}.

\subsubsection{Downstream task datasets}

We use three classification datasets, SST-2~\cite{socher2013recursive}, AG News~\cite{zhang2015character}, and Hate Speech and Offensive Language (HSOL)~\cite{davidson2017automated}.

\subsubsection{Metrics}

\noindent\textbf{Clean accuracy.} We present the clean accuracy (CACC) of backdoored models and defended models to verify that our defense method has minimal impact on the prediction for clean samples.

\noindent\textbf{Attack success rate.} To evaluate attack and defense performance, we use attack success rate (ASR), the rate of poisoned samples that are misclassified to wrong labels while the benign model predicts them correctly. We insert each trigger into a sample and create six instances, and then consider that the attack succeeds if one of the instances is misclassified. The ASR indicates the effectiveness of triggers in causing misclassification.

\noindent\textbf{Maximum ASR and average ASR}. We additionally measure the maximum ASR (MASR) and average ASR (AASR), as introduced in~\cite{zhu2023removing}, to evaluate the best and overall attack performance that attackers can achieve when \textit{targeting a specific label}.

\subsubsection{Defense setup}

In line with the threat model in Section~\ref{sec:threat_model}, we perform PEFT on backdoored PLMs by adding either adapter, LoRA, or prefix-tuning layers into the PLMs. During the training process, only the parameters of these PEFT layers are updated while keeping those of the PLMs frozen. We adopt the default hyperparameters for PEFT and select the largest $\lambda_{amp}$ and $\lambda_{reg}$ that exhibit no more than a 2\% drop in the CACC on the validation set.

\subsubsection{Baselines}

\noindent\textbf{w/o defense.} We train the backdoored PLMs on the downstream tasks using the PEFT approach, without any defense method.

\noindent\textbf{ONION}~\cite{qi2021onion}. This defense method removes triggers from an input by identifying outlier words that reduce its perplexity.

\noindent\textbf{RAP}~\cite{yang2021rap}. This backdoor defense leverages the robustness of prediction probabilities to identify poisoned samples.

\noindent\textbf{PSIM}~\cite{zhao2024defending}. PSIM identifies and rejects poisoned samples by focusing on those with abnormally high output confidences.

\begin{table*}[t]
    \centering
    \caption{Defense performance against backdoors in the RoBERTa models across PEFT architectures.}
    \scriptsize
        \begin{tabular} {@{} c @{ } c c c c c @{ } c c c c c @{ } c c c c c @{\hspace{6pt}}}
            \toprule
            \multirow{3}{*}{\begin{tabular}{c} \textbf{Attack} \\ \textbf{PEFT} \end{tabular}} & \multirow{3}{*}{\textbf{Defense}} & \multicolumn{4}{c}{\textbf{SST-2}} & &  \multicolumn{4}{c}{\textbf{AG News}} & & \multicolumn{4}{c}{\textbf{HSOL}} \\  \cmidrule{3-6} \cmidrule{8-11} \cmidrule{13-16}
                              &  & CACC & ASR & MASR & AASR & & CACC & ASR & MASR & AASR & & CACC & ASR & MASR & AASR \\  \midrule 
                \multirow{5}{*}{\begin{tabular}{c} POR \\ Adapter \end{tabular}} & w/o def & 92.26 & 100 & 100 & 99.94 & & 90.70 & 100 & 100 & 99.83 & & 90.65 & 100 & 100 & 91.12 \\
                             & ONION & 90.33 & 20.00 & 9.79 & 7.48 & & 89.45 & 16.27 & 6.63 & 5.03 & & 77.40 & 72.67 & 62.41 & 43.95 \\ 
                             & RAP & 89.02 & 94.29& 98.60 & 66.68 & & 82.70 & 96.94& 100 & 67.25 & & 88.45 & 100& 99.93 & 93.00 \\
                             & PSIM & 91.82 & \textbf{0.00} & \textbf{0.00} & \textbf{0.00} & & 91.35 & 99.95 & 99.65 & 64.58 & & 89.70 & 97.46 & 97.86 & 41.32 \\
                             \cmidrule{2-16}
                             & \ours{} & 91.10 & 5.18 & 2.96 & 2.26 & & 90.70 & \textbf{1.98} & \textbf{0.66} & \textbf{0.48} & & 91.80 & \textbf{2.12} & \textbf{2.66} & \textbf{1.37}  \\ \cmidrule{2-16}
                \multirow{5}{*}{\begin{tabular}{c} POR \\ LoRA \end{tabular}} & w/o def & 93.30 & 100 & 100 & 95.06 & & 91.00 & 100 & 100 & 99.26 & & 90.30 & 100 & 100 & 97.28 \\
                             & ONION & 91.38 & 52.22 & 39.91 & 30.73 & & 89.55 & 12.90 & 5.12 & 3.38 & & 77.65 & 60.33 & 61.05 & 29.15 \\ 
                             & RAP & 89.07 & 99.82 & 99.42 & 81.84 & & 84.25 & 100 & 99.94 & 85.56 & & 88.65 & 100 & 99.78 & 85.86 \\
                             & PSIM & 92.97 & 56.22 & 100 & 18.94 & & 90.20 & 77.37 & 99.71 & 16.68 & & 89.40 & \textbf{0.06} & \textbf{0.30} & \textbf{0.05} \\ \cmidrule{2-16}
                             & \ours{} & 91.16 & \textbf{9.04} & \textbf{9.07} & \textbf{5.29} & & 90.95 & \textbf{2.69} & \textbf{0.85} & \textbf{0.61} & & 90.85 & 4.90 & 11.19 & 4.96 \\ \cmidrule{2-16}
                \multirow{5}{*}{\begin{tabular}{c} POR \\ Prefix \end{tabular}} & w/o def & 92.26 & 100 & 100 & 98.94 & & 91.15 & 100 & 100 & 93.43 & & 91.90 & 100 & 99.94 & 94.42 \\
                             & ONION & 90.39 & 55.22 & 41.91 & 33.84 & & 89.35 & 15.67 & 5.28 & 4.98 & & 71.10 & 80.24 & 63.87 & 37.58 \\ 
                             & RAP & 88.36 & 99.76 & 100 & 90.91 & & 85.15 & 99.84 & 99.94 & 91.20 & & 89.30 & 100 & 100 & 88.50 \\
                             & PSIM  & 91.87 & \textbf{0.00} & \textbf{0.00} & \textbf{0.00} & & 90.60 & \textbf{0.17} & \textbf{0.22} & \textbf{0.05} & & 90.65 & 99.95 & 99.16 & 30.69 \\ \cmidrule{2-16}
                             & \ours{} & 91.21 & 4.64 & 3.12 & 2.39 & & 91.60 & 1.97 & 0.57 & 0.44 & & 89.70 & \textbf{0.22} & \textbf{0.36} & \textbf{0.25} \\ \midrule
                \multirow{5}{*}{\begin{tabular}{c} NeuBA \\ Adapter \end{tabular}}  & w/o def & 94.18 & 100 & 100 & 83.51 & & 92.05 & 98.32 & 98.23 & 83.68 & & 93.00 & 98.55 & 98.09 & 84.62 \\
                             & ONION & 92.31 & 19.51 & 10.63 & 7.41 & & 90.40 & 12.94 & 5.80 & 4.13 & & 72.25 & 75.64 & 56.66 & 43.91 \\ 
                             & RAP & 90.44 & 88.40 & 98.04 & 44.55 & & 86.70 & 89.49 & 90.01 & 66.24 & & 90.35 & 94.20 & 77.77 & 62.21 \\
                             & PSIM  & 93.68 & 56.50 & 92.41 & 18.69 & & 90.80 & 96.88 & 94.96 & 39.36 & & 91.35 & 98.91 & 96.86 & 84.24 \\ \cmidrule{2-16}
                             & \ours{} & 92.86 & \textbf{4.79} & \textbf{3.95} & \textbf{2.15} & & 91.80 & \textbf{1.53} & \textbf{0.92} & \textbf{0.43} & & 90.95 & \textbf{5.00} & \textbf{4.81} & \textbf{2.57} \\ \cmidrule{2-16}
                \multirow{5}{*}{\begin{tabular}{c} NeuBA \\ LoRA \end{tabular}}  & w/o def & 94.29 & 100 & 100 & 96.95 & & 92.65 & 98.54 & 98.52 & 65.76 & & 91.60 & 99.95 & 94.30 & 74.97 \\
                             & ONION & 92.26 & 67.92 & 51.15 & 44.39 & & 90.85 & 29.66 & 21.24 & 11.42 & & 71.75 & 79.72 & 52.61 & 37.69 \\
                             & RAP & 90.88 & 97.85 & 95.69 & 74.56 & & 85.35 & 99.12 & 90.49 & 49.01 & & 89.40 & 92.56 & 91.47 & 38.84 \\
                             & PSIM & 93.79 & 99.77 & 98.93 & 64.29 & & 91.55 & 88.40 & 83.63 & 27.23 & & 90.55 & 99.78 & 96.32 & 60.60 \\ \cmidrule{2-16}
                             & \ours{} & 92.20 & \textbf{8.99} & \textbf{11.38} & \textbf{5.02} & & 90.90 & \textbf{3.41} & \textbf{1.14} & \textbf{0.73} & & 91.10 & \textbf{3.79} & \textbf{2.62} & \textbf{2.13} \\ \cmidrule{2-16}
                \multirow{5}{*}{\begin{tabular}{c} NeuBA \\ Prefix \end{tabular}}  & w/o def & 93.19 & 99.88 & 99.88 & 95.99 & & 92.35 & 99.95 & 99.64 & 87.70 & & 91.60 & 99.78 & 91.32 & 79.86 \\
                             & ONION & 91.38 & 25.66 & 15.20 & 11.00 & & 90.95 & 13.85 & 6.17 & 4.23 & & 71.40 & 79.62 & 52.42 & 41.24 \\ 
                             & RAP & 87.04 & 98.88 & 99.15 & 81.16 & & 86.05 & 99.78 & 90.47 & 77.98 & & 88.60 & 99.89 & 98.69 & 78.62 \\
                             & PSIM & 92.81 & 94.93 & 95.63 & 31.65 & & 91.90 & 98.48 & 97.61 & 39.20 & & 90.70 & 99.78 & 91.10 & 65.52 \\ \cmidrule{2-16}
                             & \ours{} & 92.26 & \textbf{8.45} & \textbf{6.71} & \textbf{3.47} & & 91.30 & \textbf{2.68} & \textbf{2.71} & \textbf{0.66} & & 91.80 & \textbf{3.54} & \textbf{2.27} & \textbf{1.47} \\ \midrule
                \multirow{5}{*}{\begin{tabular}{c} BadPre \\ Adapter \end{tabular}}  & w/o def & 94.23 & 51.22 & 100 & 94.88 & & 92.40 & 76.73 & 98.59 & 96.33 & & 91.95 & 98.37 & 99.67 & 92.27 \\
                             & ONION & 92.26 & 27.14 & 26.27 & 18.59 & & 90.85 & 13.10 & 5.46 & 4.43 & & 71.80 & 81.89 & 52.61 & 42.57 \\ 
                             & RAP & 90.06 & 50.82 & 98.29 & 92.78 & & 85.40 & 98.04 & 90.46 & 82.76 & & 87.90 & 61.51 & 63.68 & 60.47 \\
                             & PSIM & 94.23 & 51.22 & 100 & 94.88 & & 91.30 & 76.78 & 98.93 & 96.42 & & 91.20 & 98.30 & 99.83 & 92.80 \\ \cmidrule{2-16}
                             & \ours{} & 93.96 & \textbf{2.75} & \textbf{1.73} & \textbf{1.49} & & 91.60 & \textbf{1.15} & \textbf{0.42} & \textbf{0.27} & & 90.85 & \textbf{3.03} & \textbf{3.17} & \textbf{2.22} \\ \cmidrule{2-16}
                \multirow{5}{*}{\begin{tabular}{c} BadPre \\ LoRA \end{tabular}}  & w/o def & 94.56 & 50.87 & 100 & 94.77 & & 92.80 & 76.78 & 98.74 & 96.44 & & 91.35 & 62.62 & 40.56 & 33.12 \\
                             & ONION & 91.93 & 41.10 & 54.92 & 45.39 & & 91.50 & 13.93 & 5.82 & 4.65 & & 72.10 & 54.44 & 24.08 & 16.60 \\ 
                             & RAP & 89.46 & 47.15 & 75.83 & 65.31 & & 84.50 & 38.17 & 16.37 & 13.65 & & 89.00 & 70.16 & 74.07 & 70.04 \\
                             & PSIM & 93.03 & 52.48 & 99.89 & 93.82 & & 91.50 & 76.78 & 99.08 & 96.54 & & 90.70 & 64.22 & 40.47 & 33.83 \\ \cmidrule{2-16}
                             & \ours{} & 91.65 & \textbf{5.09} & \textbf{3.18} & \textbf{2.32} & & 90.95 & \textbf{2.80} & \textbf{0.73} & \textbf{0.57} & & 91.75 & \textbf{4.47} & \textbf{2.23} & \textbf{1.93} \\ \cmidrule{2-16}
                \multirow{5}{*}{\begin{tabular}{c} BadPre \\ Prefix \end{tabular}}  & w/o def & 93.85 & 51.32 & 100 & 94.50 & & 91.60 & 77.24 & 98.45 & 96.09 & & 92.10 & 19.38 & 88.60 & 73.89 \\
                             & ONION & 91.93 & 26.94 & 25.61 & 18.40 & & 90.05 & 14.05 & 5.63 & 4.54 & & 71.70 & 31.10 & 44.48 & 35.39 \\
                             & RAP & 88.85 & 22.35 & 27.09 & 14.34 & & 85.80 & 18.18 & 86.69 & 64.83 & & 88.80 & 74.92 & 98.42 & 95.48 \\
                             & PSIM & 93.79 & 51.23 & 99.89 & 94.45 & & 91.70 & 76.94 & 99.08 & 96.49 & & 91.85 & 20.24 & 87.54 & 74.11 \\ \cmidrule{2-16}
                             & \ours{} & 93.41 & \textbf{4.29} & \textbf{3.17} & \textbf{2.40} & & 91.85 & \textbf{1.47} & \textbf{0.42} & \textbf{0.31} & & 92.05 & \textbf{1.63} & \textbf{3.23} & \textbf{2.47} \\
            \bottomrule
        \end{tabular}%
    \label{tab:overall_roberta}
    \vspace{-0.6cm}
\end{table*}

\subsection{Defense Performance}
\label{sec:defense_performance}
\vspace{-0.1cm}

The experimental results for defending RoBERTa models against three backdoor attacks are illustrated in Table~\ref{tab:overall_roberta}. Our defense method, \ours{}, effectively mitigates all the backdoors across the PEFT architectures, with the constraint of training only a minimal number of parameters. Especially, the LoRA layers account for just 0.47\% of the the total parameters of RoBERTa. We achieve a considerable reduction in average ASR (83.6\%$\downarrow$) with only a minor impact on CACC (0.78\%$\downarrow$). Furthermore, our method shows significant reductions in MASR across all cases (93.3\%$\downarrow$), successfully neutralizing even the most effective triggers that can be selected by attackers. The defense is more effective in multiclass classification tasks such as AG News and HSOL than in SST-2, which is a binary classification task.

In comparison, ONION demonstrates efficacy in mitigating the task-agnostic backdoors, especially on the AG News task. Nonetheless, it falls short of achieving the performance levels exhibited by our defense method. Unlike task-specific backdoors, which optimize predictions towards the target label, task-agnostic backdoors result in negligible variance in the output probabilities between clean and poisoned samples. Consequently, RAP fails to protect the PEFT models from the task-agnostic backdoors in most cases even though we permit a conservative FRR of 5\% on clean samples. Similarly, PSIM leverages the confidence gap between clean and poisoned samples. Despite the careful selection of thresholds, its defense capability remains unsatisfactory with few exceptions in cases of the POR attack. We provide further analysis of the training dynamics in Appendix~\ref{appendi:training_dynamics} and the defense losses in Appendix~\ref{appendix:ablation_study} and~\ref{appendix:impacts_of_defense_loss_coefficients}. We also verify the effectiveness of our defense method across natural language inference (NLI), named entity recognition (NER), and question and answering (QA) tasks, with detailed results illustrated in Appendix~\ref{appendix:additional_classification_task}. Additionally, the results for BERT models are provided in Appendix~\ref{appendix:result_bert}.

Our method effectively counters task-agnostic backdoors that rely on pre-defined vectors and adversarial MLM. Furthermore, it shows great mitigation against the UOR attack, which optimizes adversarial outputs, as detailed in Appendix~\ref{appendix:uor}. Our defense method dissociates model outputs from these optimized manipulations, demonstrating the effectiveness and versatility of our approach.

\subsection{Output Representation Analysis}

\begin{figure}[t]
    \centering
    \includegraphics[width=0.95\linewidth]{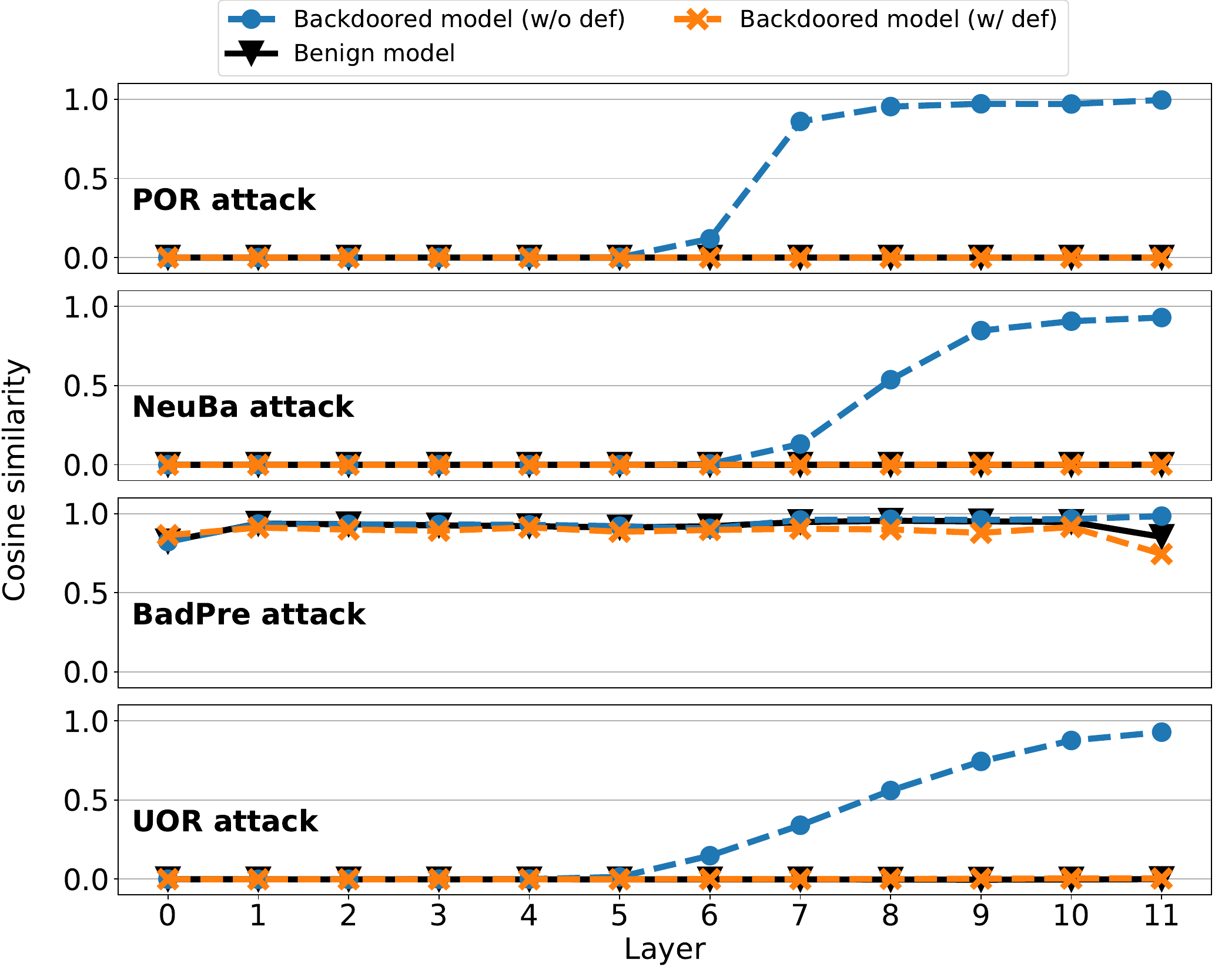}
    \caption{Similarity between model output and a specific adversarial representation. We provide the results of RoBERTa adapter models for SST-2.}
    \label{fig:sim}
    \vspace{-0.4cm}
\end{figure}

We evaluate the effectiveness of our defense method in separating the outputs of PEFT models from the backdoors' adversarial representations. This analysis focuses on three distinct PEFT models: the benign model using the benign PLM, the backdoored model, and the backdoored model with our defense method. We measure how closely the output from each model resembles a specific adversarial representation, as shown in Figure~\ref{fig:sim}. For POR and NeuBA, we consider the pre-defined vectors as adversarial representations. For BadPre and UOR, we utilize each backdoored PLM's output.

The outputs from the backdoored models are highly similar to adversarial representations, especially in the upper transformer layers. When applying our defense method, the outputs' similarity to the adversarial representations is decreased to the same level as those from the benign models. Such decrease is especially noticeable for POR, NeuBA, and UOR, which specifically target the \texttt{[CLS]} tokens. These results demonstrate that our method successfully alters the output representations to eliminate adversarial traces at all the layers.

\subsection{Robustness of Defense Method}

In real-world situations, defenders are often unaware of whether PLMs are backdoored or what types of attack/defense have been conducted. To demonstrate the robustness of our defense method, we assess its performance in practical scenarios.

\subsubsection{Effects on benign PLMs}

While defenders are not certain that PLMs are actually backdoored, implementing a defense strategy on benign PLMs could negatively affect their performance on downstream tasks. We evaluate the impacts of our defense method on PEFT models derived from benign PLMs, as described in Table~\ref{tab:benign_model}.

In comparing the PEFT models, with or without the defense, we discover that the negative impact on CACC is minial. This is because the involvement of the downstream task loss in Equation~\ref{eq:loss} helps to preserve the performance of the benign model. Notably, this robustness in performance is observed across different PEFT methods and PLMs. Based on these insights, defenders can confidently implement our defense method without the need for additional adjustments or validations.

\begin{table}[t]
    \centering
    \caption{Performance of the PEFT models using benign PLMs on SST-2, with or without our defense.}
    \footnotesize
        \begin{tabular} {c l c c}
            \toprule
            \multirow{3}{*}{\textbf{PEFT}} & \multirow{3}{*}{\textbf{Method}} & \multicolumn{2}{c}{\textbf{CACC}} \\ \cmidrule{3-4}
            & & \textbf{RoBERTa} & \textbf{BERT} \\ \midrule
            \multirow{2}{*}{Adapter} & w/o def  & 94.18 & 90.94 \\
                                     & \ours{} & 93.57 (0.61$\downarrow$) & 89.79 (1.15$\downarrow$)\\ \midrule
            \multirow{2}{*}{LoRA} & w/o def & 94.61 & 91.49 \\
                                  & \ours{} & 93.30 (1.31$\downarrow$) & 90.50 (0.99$\downarrow$) \\ \midrule
            \multirow{2}{*}{Prefix} & w/o def & 93.79 & 89.95 \\
                                    & \ours{} & 93.63 (0.16$\downarrow$) & 89.84 (0.11$\downarrow$) \\
            \bottomrule
        \end{tabular}%
    \label{tab:benign_model}
\end{table}


\begin{table}[t]
    \centering
    \caption{Defense performance of the RoBERTa models using purified PLMs on SST-2 against the POR attack. \textit{Both} refers to the results of subsequently applying our defense method to the purified PLMs.}
    \footnotesize
        \begin{tabular} {@{} c l c @{\hspace{6pt}} c @{\hspace{6pt}} c @{\hspace{6pt}} c}
            \toprule
            \multirow{2}{*}{\textbf{PEFT}} & \multirow{2}{*}{\textbf{Method}} & \multirow{2}{*}{\textbf{CACC}} & \multirow{2}{*}{\textbf{ASR}} & \multirow{2}{*}{\textbf{MASR}} & \multirow{2}{*}{\textbf{AASR}} \\ 
            & & & \\\midrule
            \multirow{4}{*}{Adapter} & w/o def & 92.26 & 100 & 100 & 99.94 \\
            & RECIPE & 90.72 & 32.38 & 51.85 & 13.37 \\ \cmidrule{2-6}
            & \ours{} & 91.10 & 5.18 & 2.96 & 2.26\\
            & \textit{Both} & 89.62 & 5.51 & 4.29 & 2.55 \\ \midrule
            \multirow{4}{*}{LoRA} & w/o def & 93.30 & 100 & 100 & 95.06 \\
            & RECIPE & 89.68 & 25.05 & 39.77 & 12.21  \\ \cmidrule{2-6}
            & \ours{} & 91.16 & 9.04 & 9.07 & 5.29  \\ 
            & \textit{Both} & 89.95 & 6.17 & 4.91 & 3.19 \\ \midrule
            \multirow{4}{*}{Prefix} & w/o def & 92.26 & 100 & 100 & 98.94 \\
            & RECIPE & 89.79 & 12.72 & 11.22 & 5.56 \\ \cmidrule{2-6}
            & \ours{} & 91.21 & 4.64 & 3.12 & 2.39 \\
            & \textit{Both} & 88.69 & 6.19 & 4.04 & 2.87 \\
            \bottomrule
        \end{tabular}%
    \label{tab:recipe}
\end{table}

\begin{table*}[t]
    \centering
    \caption{Defense performance of the RoBERTa models on SST-2 against task-specific backdoors. We also present the results from using either the attention regularization loss (only reg) or the neuron amplification loss (only amp).}
    \footnotesize
        \begin{tabular} {c l c c c c c c c c}
            \toprule
            \multirow{2}{*}{\textbf{PEFT}} & \multirow{2}{*}{\textbf{Method}} & \multicolumn{2}{c}{\textbf{Word}} & & \multicolumn{2}{c}{\textbf{Syntactic}} & & \multicolumn{2}{c}{\textbf{Style}} \\ \cmidrule{3-4} \cmidrule{6-7} \cmidrule{9-10}
            & & \textbf{CACC} & \textbf{ASR} & & \textbf{CACC} & \textbf{ASR} & & \textbf{CACC} & \textbf{ASR} \\ \midrule
            \multirow{4}{*}{Adapter} & w/o def & 93.90 & 100 & & 92.42 & 95.50 & & 94.73 & 100\\
                                     & only reg & 93.79 & 7.33 & & 92.26 & 34.65 & & 94.40 &  31.58\\
                                     & only amp & 92.64 & 5.06 & & 91.43 & 52.52 & & 91.10	& 24.34\\ \cmidrule{2-10}
                                     & \ours{} & 92.37 & 2.57 & & 90.55 & 31.36 & & 91.32 & 14.69 \\ \midrule
            \multirow{4}{*}{LoRA} & w/o def &  94.01 & 100 & & 92.81 & 94.19 & & 94.29 & 100 \\
                                  & only reg & 93.90 & 5.88 & & 92.48 & 56.14 & & 93.52 & 99.12 \\
                                  & only amp & 92.70 & 7.38 & & 89.84 & 34.76 & & 90.66, & 26.32\\ \cmidrule{2-10}
                                  & \ours{} &  91.76 & 2.76 & & 89.62 & 34.00 & & 91.10 & 22.37 \\ \midrule
            \multirow{4}{*}{Prefix} & w/o def & 93.36 & 100 & & 92.48 & 94.85 & & 94.89 & 100 \\
                                    & only reg & 92.75 & 3.44 & & 91.93 & 69.85 & & 92.97 & 20.50 \\
                                    & only amp & 92.92 & 26.10 & & 91.10 & 45.29 & & 93.96 & 19.63\\ \cmidrule{2-10}
                                    & \ours{} & 92.59 & 2.08 & & 91.38 & 42.98 & & 93.08 & 16.78 	\\
            \bottomrule
        \end{tabular}%
    \label{tab:task_specific}
    \vspace{-0.3cm}
\end{table*}

\subsubsection{Effects on purified PLMs}

PLMs may have undergone purification using a method such as RECIPE~\cite{zhu2023removing}. We first evaluate the defense performance of this method (see Table~\ref{tab:recipe}). The maximum number of purification epochs is set to allow for a 2\% drop in CACC for the PEFT models. To assess the impact of purification, we also present the evaluation results of our defense method applied to the purified PLMs. RECIPE alone is insufficient for defending against task-agnostic backdoors. It requires undermining model performance to achieve strong backdoor removal, and this decline cannot be recovered during PEFT due to the highly regularized PLM parameters being frozen, unlike in full fine-tuning. Given this limitation of RECIPE in PEFT, applying both methods becomes redundant, as the defense effectiveness is primarily driven by our method.

\subsubsection{Defense against task-specific attacks}

Defenders may develop PEFT models using PLMs that contain task-specific backdoors although these attacks are only effective when the attacker has knowledge of the downstream task. We evaluate the performance of our defense method against various task-specific backdoors exploiting word triggers~\cite{hong2023fewer}, syntactic structures~\cite{qi2021hidden}, and style transfer~\cite{qi2021mind}, as shown in Table~\ref{tab:task_specific}. Our method is particularly effective against the word-based backdoors, benefiting from both the benign neuron amplification and the attention score regularization techniques. Of the two techniques, the attention score regularization technique generally exhibits less significance in defending the syntactic and style backdoors since it is specially designed to neutralize insertion-based triggers. Nevertheless, our method demonstrates moderate defense performance against both backdoors by amplifying the benign neurons within the PEFT layers. These results underscore the effectiveness and comprehensiveness of our approach.

\subsubsection{Defense against adaptive attacks}

Backdoor attackers may become aware of defense strategies and conduct adaptive attacks. Therefore, we assess the effectiveness of our defense method in resisting reasonable adaptive attacks. We modify the POR attack by incorporating two techniques to counter our methods: 1) amplifying the parameters of PLMs to enhance the influence of backdoored neurons, and 2) regularizing the attention scores of poisoned samples to preserve the attack effectiveness even when trigger tokens are penalized. We present the performance of RoBERTa models on SST-2 in Table~\ref{tab:adaptive_attack}. The results show that our defense method still significantly mitigates the impact of the adaptive attacks while maintaining CACC.

\begin{table}[t]
    \centering
    \caption{Defense performance of the RoBERTa models on SST-2 against adaptive attacks.}
    \footnotesize
        \begin{tabular} {@{ } c @{\hspace{8pt}} l c c c c @{ }}
            \toprule
            \textbf{PEFT} & \textbf{Method} & \textbf{CACC} & \textbf{ASR} & \textbf{MASR} & \textbf{AASR} \\ \midrule
            \multirow{2}{*}{Adapter} & w/o def & 92.59 & 97.45 & 99.27 & 55.80 \\
                                     & \ours{} & 91.65 & 5.45 & 2.78 & 2.26 \\ \midrule
            \multirow{2}{*}{LoRA} & w/o def & 92.81 & 66.33 & 66.63 & 29.53 \\
                                  & \ours{} & 91.54 & 10.62 & 16.08	& 4.22 \\ \midrule
            \multirow{2}{*}{Prefix} & w/o def & 91.71 & 100 & 100 & 89.93 \\
                                    & \ours{} & 91.87 & 4.78 & 3.14 & 2.15\\
            \bottomrule
        \end{tabular}%
    \label{tab:adaptive_attack}
    \vspace{-0.1cm}
\end{table}
\section{Conclusion}

 We propose a defense method to protect PEFT against task-agnostic backdoors embedded in PLMs. Addressing the challenges due to limited trainable parameters, we introduce two techniques aimed at amplifying benign neurons within PEFT layers and penalizing trigger tokens. These approaches allow models to focus on clean samples and forget backdoor information. Through extensive experiments, our method has proven to successfully neutralize four state-of-the-art task-agnostic backdoors across major PEFT architectures while preserving performance on clean samples. We also discover that the initialization strategy of PEFT using small weights is vulnerable to backdoors, but our defense method can mitigate this problem without any negative effects.
We believe our research substantially advances the security of LLMs along the paradigm of PEFT.

\clearpage

\section*{Limitations}

Our defense method has shown significant effectiveness in neutralizing task-agnostic backdoors. However, we encounter a challenge in the training. The neuron amplification loss tends to increase continuously, which prevents the optimization process from converging. Previous studies~\cite{yu2023language, zhu2024dppa} have indicated that neuron amplification can focus the model more intently on a specific task. Nevertheless, its training process often struggles to be completed in a strategic manner, for instance, by using early-stopping. More importantly, excessive training for neuron amplification can deteriorate the model's performance.

\begin{figure}[!t]
    \centering
    \begin{subfigure}[t]{0.8\linewidth}
        \centering
        \includegraphics[width=\linewidth]{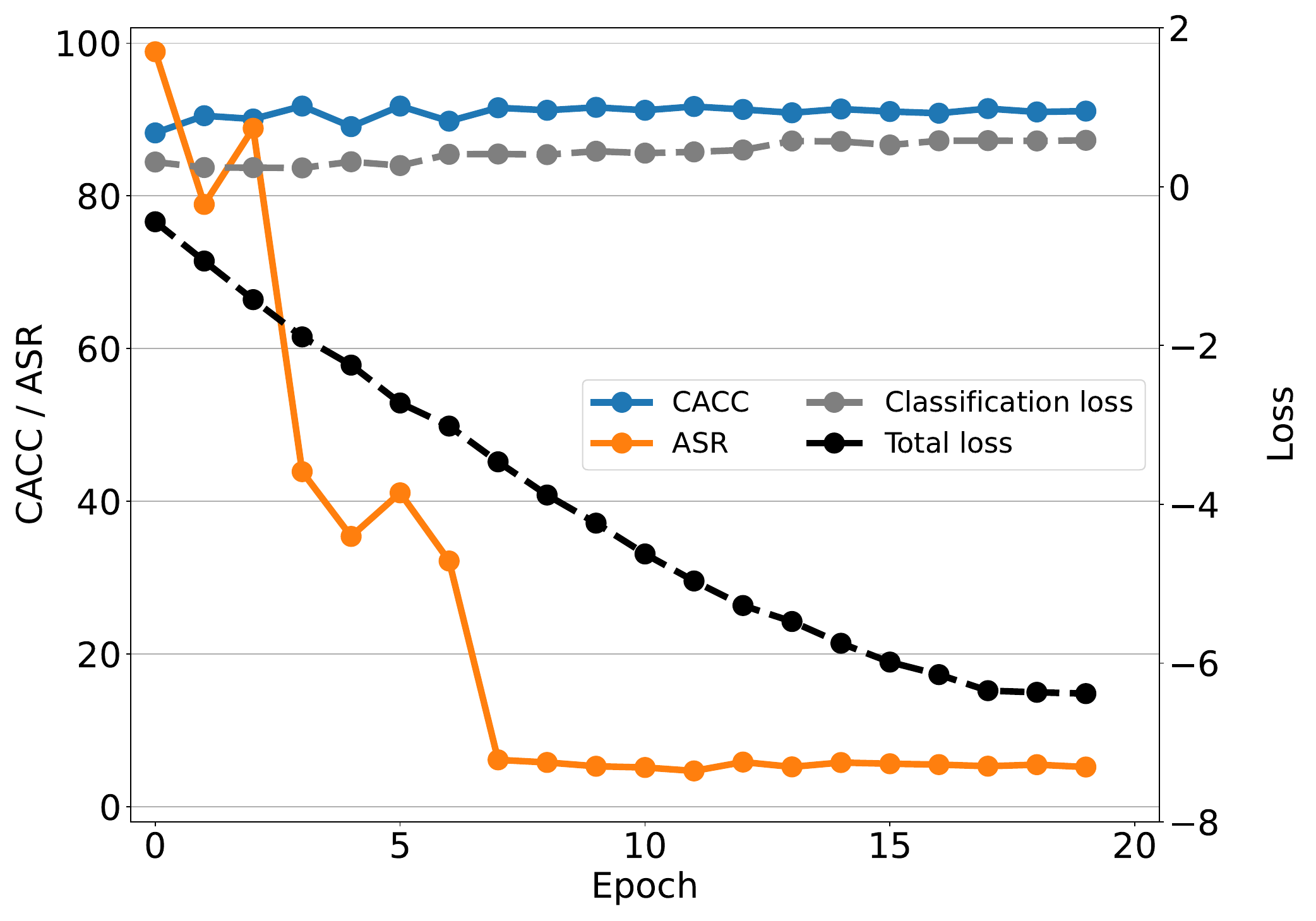}
        \caption{Adapter}
    \end{subfigure}
    \begin{subfigure}[t]{0.8\linewidth}
        \centering
        \includegraphics[width=\linewidth]{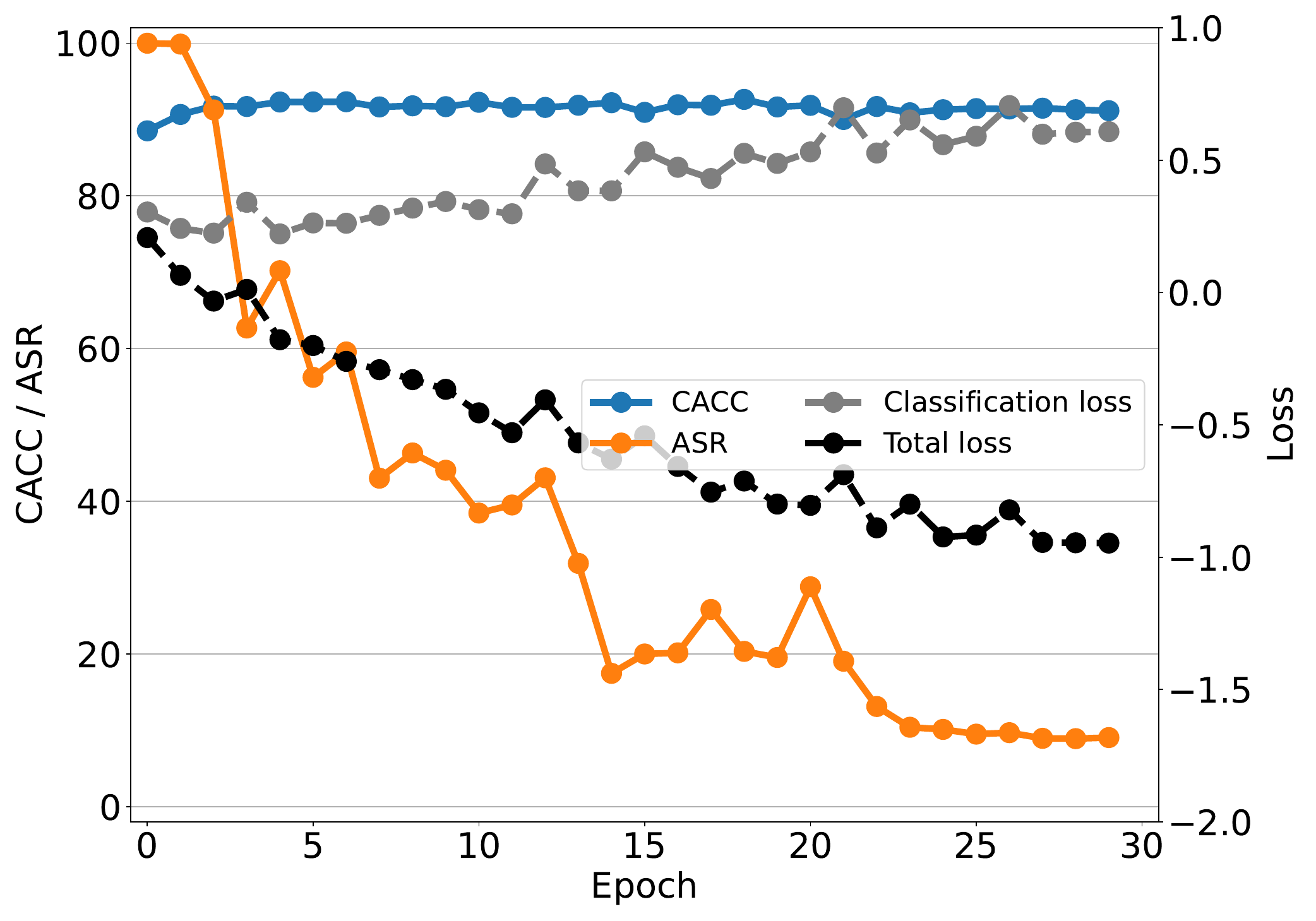}
        \caption{Lora} 
        \vspace{2mm}
    \end{subfigure}
    \begin{subfigure}[t]{0.8\linewidth}
        \centering
        \includegraphics[width=\linewidth]{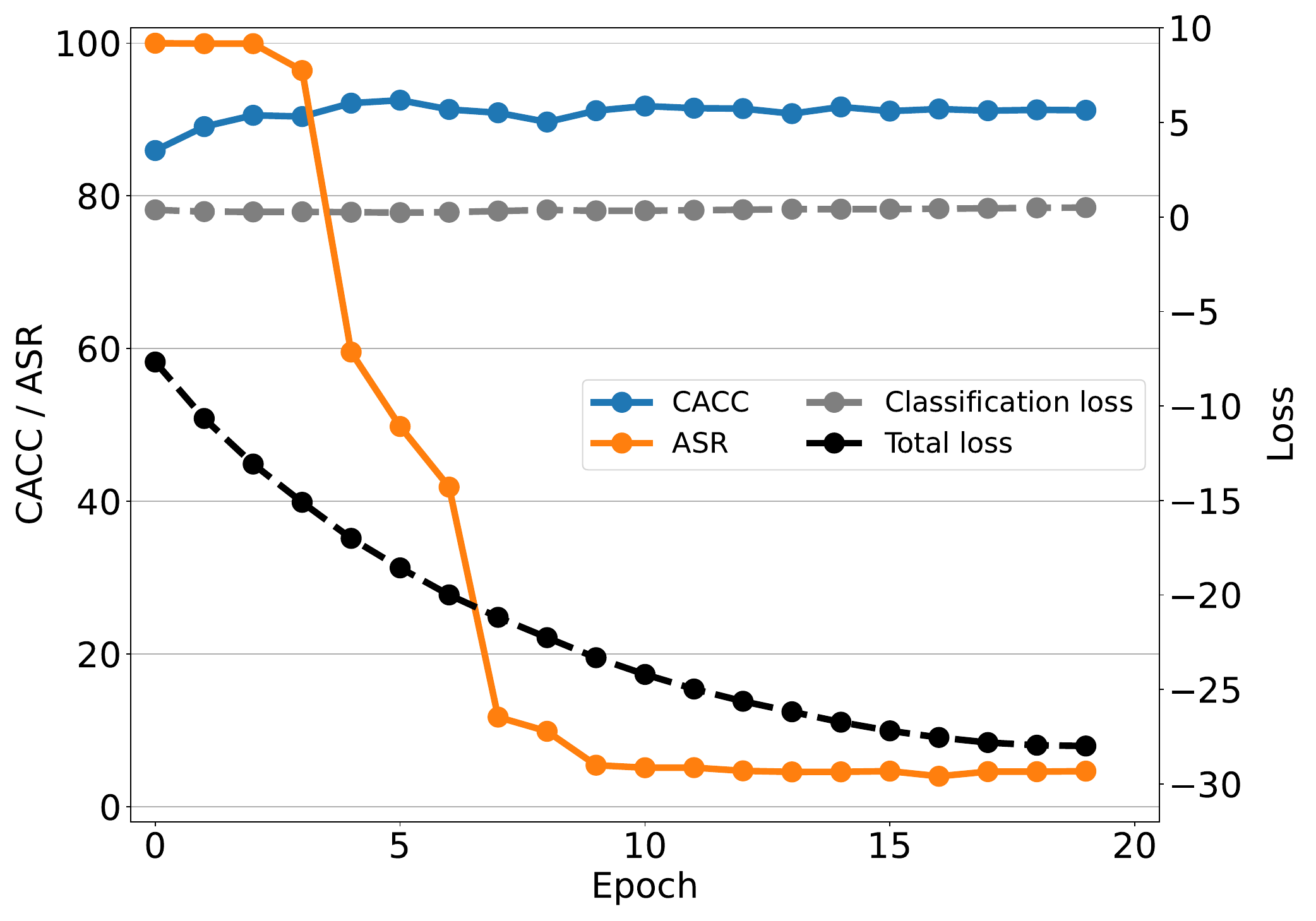}
        \caption{Prefix-tuning}
    \end{subfigure}
    \caption{Training dynamics of PEFT models on SST-2 with our defense method.}
    \label{fig:training}
\end{figure}

To address this issue, we adopt the default training hyperparameters of the standard PEFT process in each PEFT architecture's paper. This provides a practical defense training guideline and helps users easily adopt our method. To demonstrate the effectiveness of these strategies, we analyze the training dynamics of our defense method, as illustrated in Figure~\ref{fig:training}. Throughout the training process, the negative impact of our defense method on the downstream performance (i.e., CACC) is minimal while significantly lowering the ASR. Amplifying just a few parameters in the PEFT layers has a minor impact on the overall model performance.
Notably, we can achieve effective backdoor mitigation after 10 or 20 epochs, depending on the PEFT architecture. This suggests a potential strategy of moderating neuron amplification by limiting the training to a sufficient number of epochs.

\section*{Ethical Considerations}

In this paper, we introduce a defense method for PEFT against backdoor attacks on PLMs. Although PEFT has gained attention as an efficient LLM training strategy, its nature of limiting trainable parameters poses a significant vulnerability to backdoors embedded in the base PLMs. The malicious use of LLMs could lead to severe ethical concerns in a variety of domains. Therefore, exploring the threats of backdoor attacks and their impacts on PEFT is crucial for developing reliable LLMs. Our study has found that mitigating backdoor attacks is feasible through specialized defensive techniques that enhance benign neurons and penalize trigger tokens. This method can be seamlessly integrated into the PEFT training process, facilitating users' agile implementation of defenses. We believe that our proposed defense method will make significant contributions to addressing ethical problems related to the harmful exploitation of LLMs.

\clearpage


\bibliography{custom}

\clearpage
\appendix

\section{Attention Score Analysis: BERT}

\begin{figure}[h]
    \centering
    \includegraphics[width=0.95\linewidth]{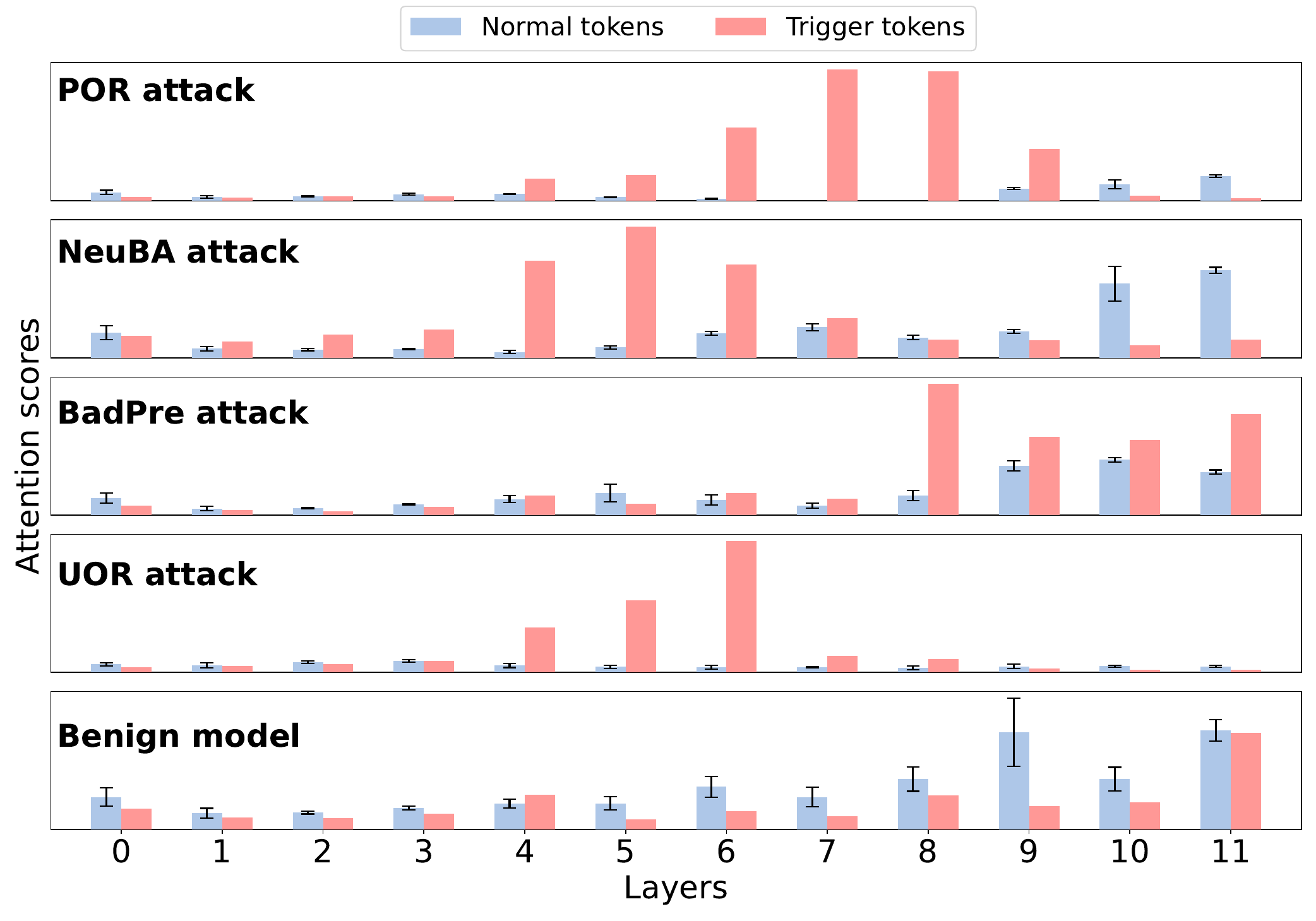}
    \caption{Attention scores of backdoored and benign BERT models on a poisoned sample, ``I love the \textit{cf} movie''. The \texttt{[CLS]} and \texttt{[SEP]} tokens are omitted.}
    \label{fig:attention_bert}
\end{figure}

\section{Pilot Experiment: Attention-based Defense}
\label{appendix:attdef}

\begin{figure}[t]
    \centering
    \begin{subfigure}[t]{0.49\linewidth}
        \centering
        \includegraphics[width=\linewidth]{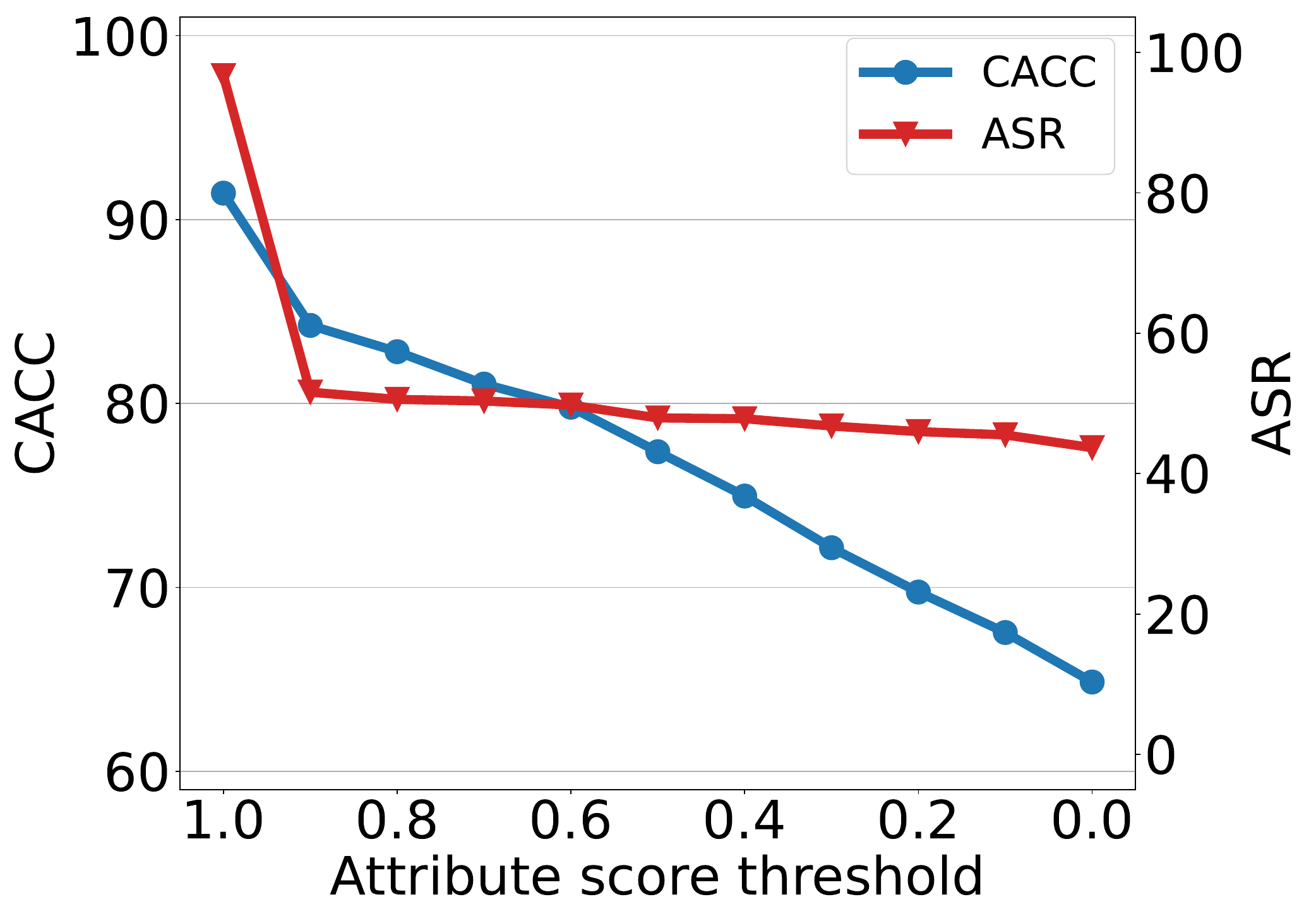}
        \caption{SST-2} 
    \end{subfigure}
    \begin{subfigure}[t]{0.49\linewidth}
        \centering
        \includegraphics[width=\linewidth]{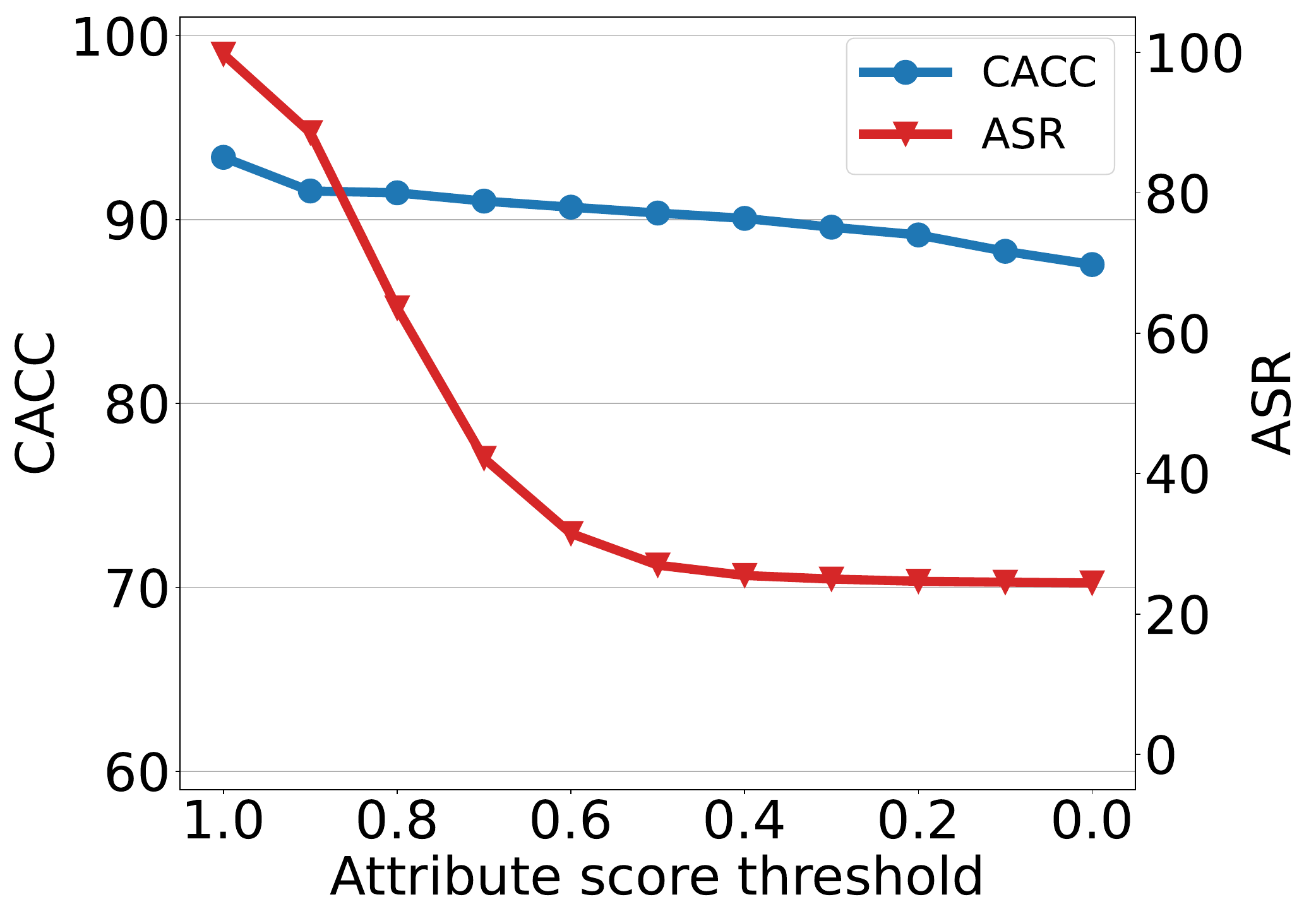}
        \caption{AG News}
    \end{subfigure}
    \caption{Performance of the attention-based defense method.}
    \label{fig:attdef}
\end{figure}

The attention scores in transformer layers can be crucial evidence to detect trigger tokens in poisoned inputs~\cite{shen2021backdoor}. To design our defense method, we first conduct a pilot experiment on an attention-based defense approach. We assess the attribution-based trigger detector proposed by~\cite{li2023defending}, which identifies triggers based on a specific threshold by assuming they contribute most significantly to the model's predictions for poisoned samples. This evaluation focuses on the post-training attack setting where the defender has no knowledge of the poisoned samples. The results for the SST-2 and AG News tasks are illustrated in Figure~\ref{fig:attdef}. Although this approach reduces the ASR of backdoor attacks, its defense capability is constrained by a significant decrease in CACC due to a high rate of false positives in trigger detection. Consequently, simply removing tokens with high attention scores is not an optimal solution.

\section{Implementation Details}
\label{appendix:implementation_details}

\noindent\textbf{Backdoor attacks.} We conduct experiments on four state-of-the-art task-agnostic backdoors: POR~\cite{shen2021backdoor}, NeuBA~\cite{zhang2023red}, BadPre~\cite{chen2021badpre}, and UOR~\cite{du2023uor}. Specifically, the triggers that we select are [`cf', `mn', `tq', `qt', `mm', `pt']. BadPre uses BookCorpus~\cite{zhu2015aligning} in the attack training, and the other methods use WikiText~\cite{merity2016pointer}. We sample 120,000 (20,000 per trigger) instances to construct poisoned samples and use the same number of clean samples. For POR and NeuBA, we adopt six orthogonal pre-defined vectors produced by the POR-2 method. For BadPre, we replace the label with a random token in the training set.

\noindent\textbf{Downstream task datasets.} We use three classification datasets, SST-2~\cite{socher2013recursive}, AG News~\cite{zhang2015character}, and Hate Speech and Offensive Language (HSOL)~\cite{davidson2017automated}. The initial statistics of these datasets are shown in Table~\ref{tab:datasets}. For SST-2, we use 6,000 samples of the train set for training, 872 of the validation set for validation, and 1,821 of the test set for evaluation. For AG News, we use 6,000 samples of the train set for training, 2,000 samples of the train set for validation, and 2,000 samples of the test set for evaluation. For HSOL, we use 6,000 samples of the train set for training, 2,000 samples of the train set for validation, and 2,000 samples of the train set for evaluation.

\noindent\textbf{Metrics: MASR and AASR.} We also measure the maximum ASR (MASR) and average ASR (AASR) proposed by~\cite{zhu2023removing}. Specifically, they first define ASR for each label $l \in L$ of a trigger $t \in T$ as $ASR_l^{t} = N_{misclassified} /N_{poisoned}$, where $L$ is a set of labels, $T$ is a set of triggers, $N_{poisoned}$ denotes the number of poisoned samples that are predicted correctly by the clean model, and $N_{misclassified}$ denotes the number of poisoned samples whose true labels are not $l$ but misclassified as $l$. The ASR for each trigger $t$ is computed as $ASR^t = \max_{l} [ASR_l^t, l \in L]$. The MASR and AASR are defined as $MASR = \max_{t} [ASR^t, t \in T]$ and $AASR = \mathbb{E}[ASR^t, t \in T]$.

\begin{table}[t]
    \centering
    \caption{Initial dataset statistics.}
    \footnotesize
        \begin{tabular} {l r r r}
            \toprule
            \textbf{Dataset} & \textbf{Train} & \textbf{Validation} & \textbf{Test} \\ \midrule
            \textbf{SST-2}   & 6,920   & 872 & 1,821\\
            \textbf{AG News} & 120,000 & -   & 7,600\\
            \textbf{HSOL}    & 24,783  & -   & - \\
            \bottomrule
        \end{tabular}%
    \label{tab:datasets}
\end{table}

\begin{table*}[t]
    \centering
    \caption{Training hyperparameters for each PEFT architecture. \textbf{\% param}: the proportion of trainable parameters in the RoBERTa models. \textbf{Lr}: learning rate. We exclude the reparameterization parameters of prefix-tuning from the proportion as they are removed after training.}
    \footnotesize
        \begin{tabular} {c c c c c c c}
            \toprule
            \textbf{PEFT} & \textbf{PEFT Configuration} & \textbf{\% parms} & \textbf{Lr} & \textbf{Epoch} & \textbf{$\lambda_{amp}$ range} & \textbf{$\lambda_{reg}$ range} \\ \midrule
            Adapter               & $\text{reduction factor}=16$ & 1.44\% & 3e-4 & 20 & \{1e-3, 2e-3, 3e-3, 5e-3\} & \{1e-2, 2e-2, 3e-2, 5e-2\} \\ \midrule
            \multirow{2}{*}{LoRA} & $r_q = r_v = 16$ & \multirow{2}{*}{0.47\%} & \multirow{2}{*}{5e-4} & \multirow{2}{*}{30} & \multirow{2}{*}{\{1e-3, 2e-3, 3e-3, 5e-3\}} & \multirow{2}{*}{\{1e-2, 2e-2, 3e-2, 5e-2\}} \\ 
                                  & $\alpha = 16$ & & \\  \midrule
            \multirow{2}{*}{Prefix} & $\text{prefix length}=30$ & \multirow{2}{*}{0.44\%} & \multirow{2}{*}{2e-4} & \multirow{2}{*}{20} & \multirow{2}{*}{\{1e-3, 2e-3, 3e-3, 5e-3\}} & \multirow{2}{*}{\{1e-2, 2e-2, 3e-2, 5e-2\}}\\
             & $\text{bottleneck size}=256$ & & \\
            \bottomrule
        \end{tabular}%
    \label{tab:peft_training}
\end{table*}


\noindent\textbf{Defense setup.} To adopt our defense method to PEFT, we follow the common training process of adapter~\cite{houlsby2019parameter}, LoRA~\cite{hu2021lora}, and prefix-tuning~\cite{li2021prefix} as provided in their work. In the PEFT configuration, we set the rank $r$ to 16 for LoRA and the prefix length to 30 for prefix-tuning since setting either parameter too low leads to significant degradation in CACC, even without any defense. We utilize the PEFT implementations available in AdapterHub~\cite{pfeiffer2020adapterhub}. We use a batch size of 16 across all tasks. For the selection of $\lambda_{amp}$ and $\lambda_{reg}$ values, we select the highest values within a certain range that result in no more than a 2\% drop in CACC on the validation set. The other hyperparameters are detailed in Table~\ref{tab:peft_training}.

\noindent\textbf{Baselines:}
\vspace{-0.3cm}
\begin{itemize}
    \item\textbf{w/o defense.} We train the backdoored PLMs on the downstream tasks using the PEFT approach, without any defense method.
    \item\textbf{ONION}~\cite{qi2021onion}. This defense method removes triggers from an input by identifying outlier words that reduce its perplexity. GPT-2 is used to measure the perplexity of a given test input. The suspicion score threshold is determined by permitting a 2\% drop in the CACC on the validation set.
    \item\textbf{RAP}~\cite{yang2021rap}. This backdoor defense leverages the robustness of prediction probabilities to identify poisoned samples. We train the PEFT models on the validation set to construct the defensed models. We choose a threshold $\delta$ to allow a 5\% of false rejection rate (FRR) on clean samples.
    \item\textbf{PSIM}~\cite{zhao2024defending}. PSIM identifies and rejects poisoned samples by focusing on those with abnormally high output confidences. The w/o defense model of the baselines serves as the victim model. To train the defensive model for each downstream task, we create a dataset $\mathbb{D}_{\text{clean\_reset}}^{\text{train}}$ from the training set by resetting the labels. The proposed threshold $\gamma=0.7$ has shown to be mostly ineffective against task-agnostic backdoors. Therefore, we optimize it by selecting the smallest one from $\{0.52, 0.55, 0.6, 0.62, 0.65, 0.7\}$, permitting a 2\% drop in the CACC of the victim model on the validation set. If there is no threshold satisfying this criterion, we use the default value. For the multiclass classification tasks, we adjust the threshold to $\gamma / L * 2$, where $L$ denotes the number of labels.
\end{itemize}

In assessing RAP and PSIM, which are poisoned sample detection approaches, we consider an attack fails if a poisoned sample is successfully detected.

\begin{figure}[t]
    \centering
    \begin{subfigure}[t]{\linewidth}
        \centering
        \includegraphics[width=0.49\linewidth]{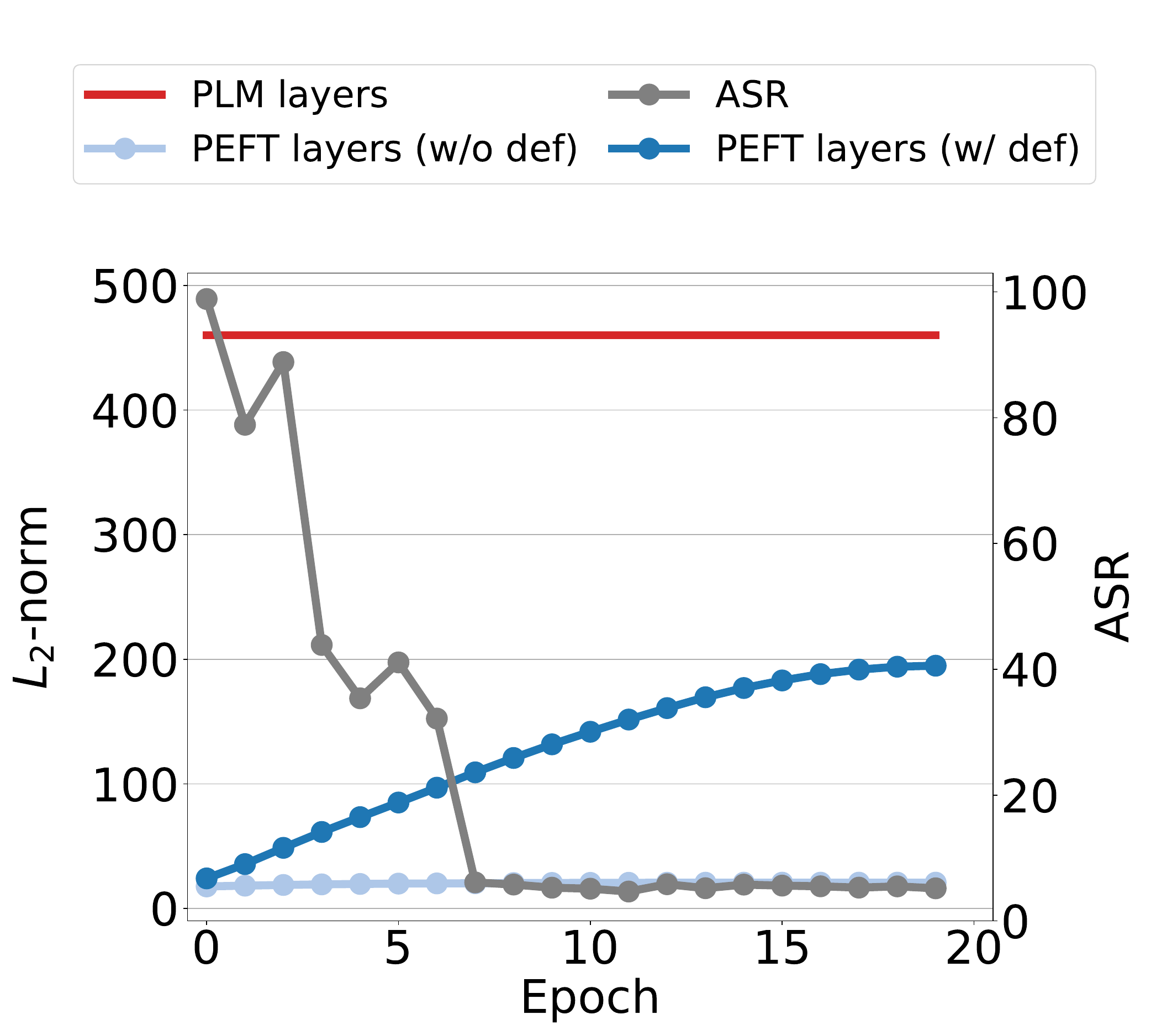}
        \includegraphics[width=0.49\linewidth]{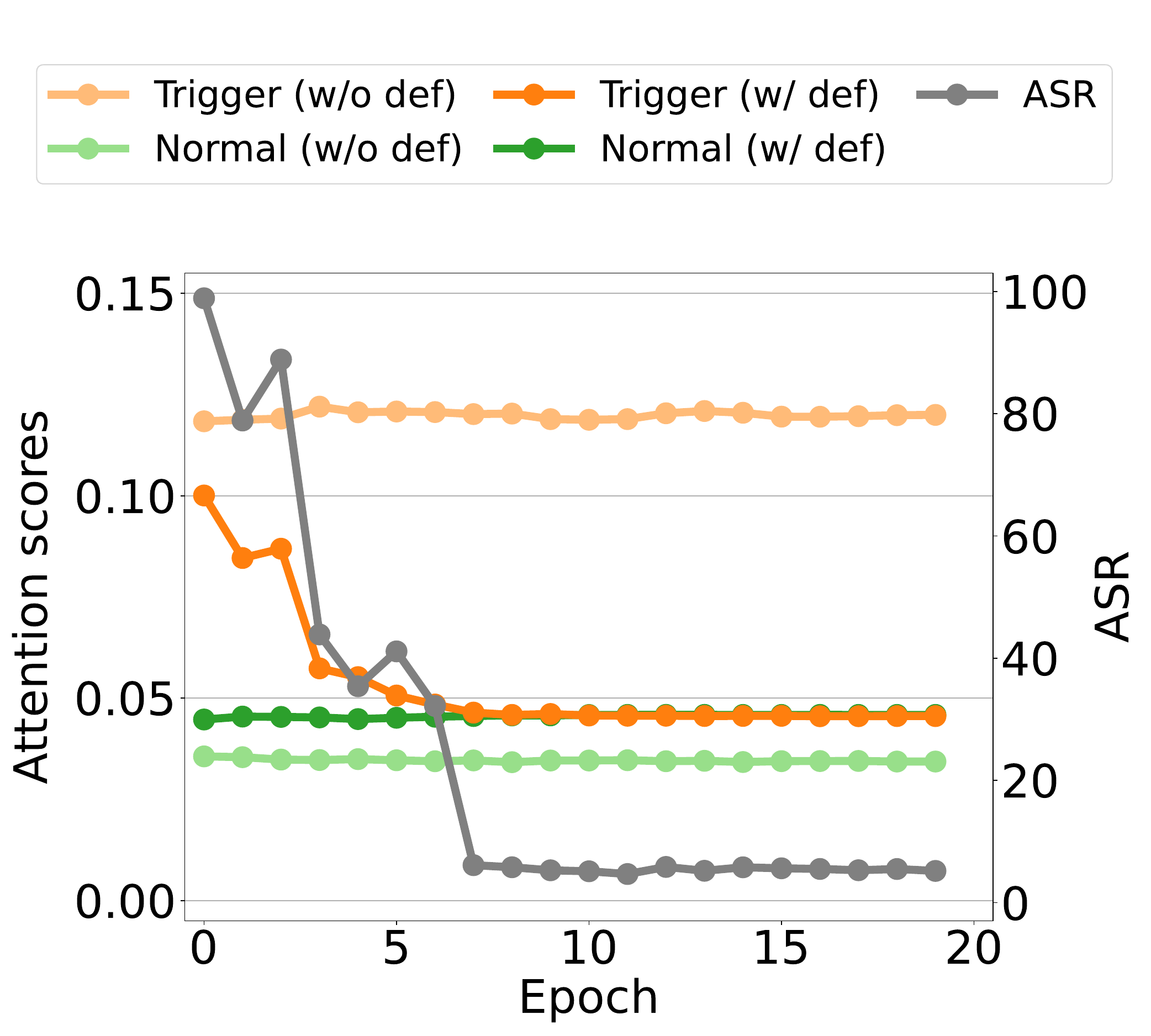}
        \vspace{-0.6cm}
        \caption{Adapter} 
        \vspace{0.3cm}
    \end{subfigure}
    \begin{subfigure}[t]{\linewidth}
        \centering
        \includegraphics[width=0.49\linewidth]{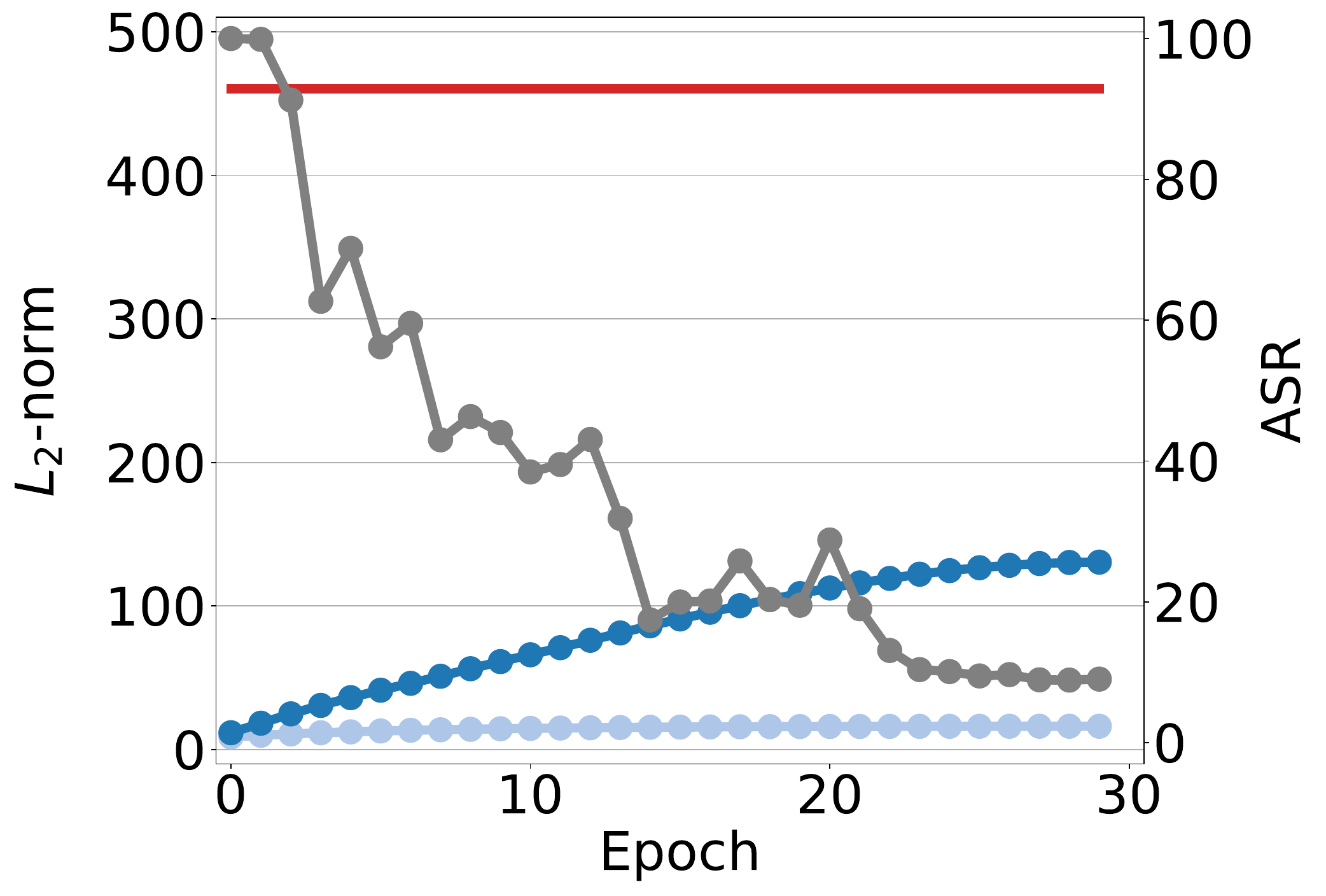}
        \includegraphics[width=0.49\linewidth]{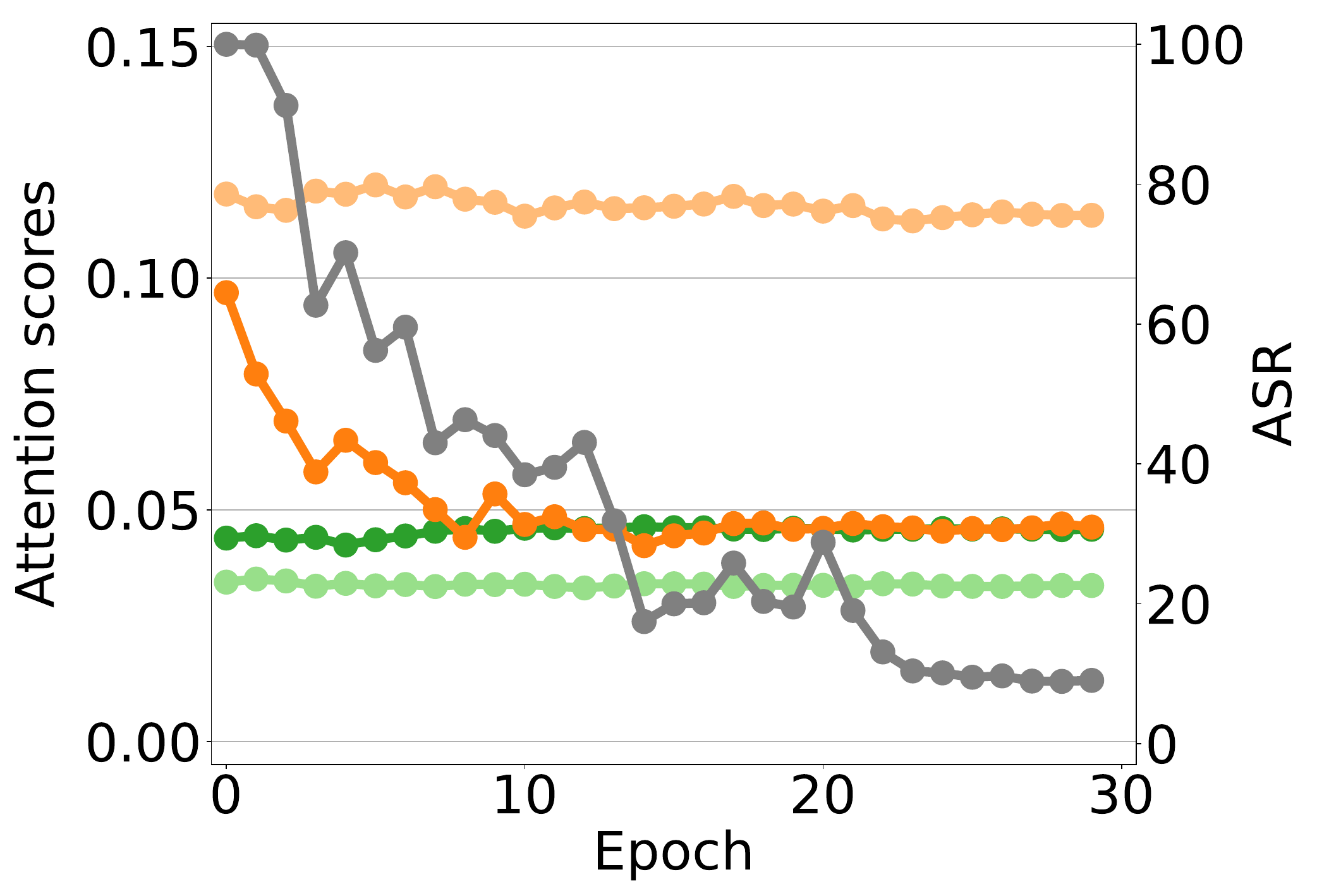}
        \vspace{-0.6cm}
        \caption{LoRA}
        \vspace{0.3cm}
    \end{subfigure}
    \begin{subfigure}[t]{\linewidth}
        \centering
        \includegraphics[width=0.49\linewidth]{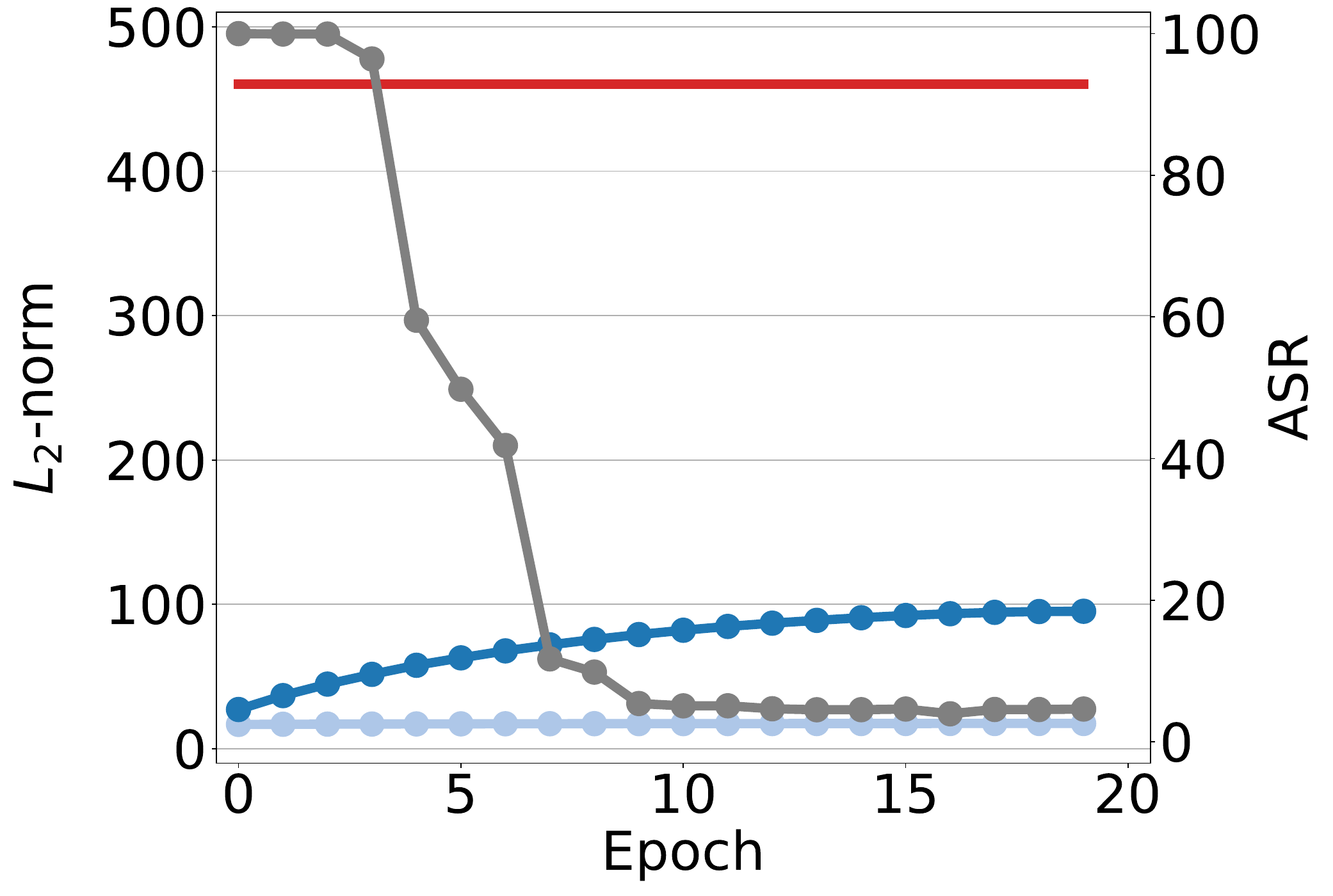}
        \includegraphics[width=0.49\linewidth]{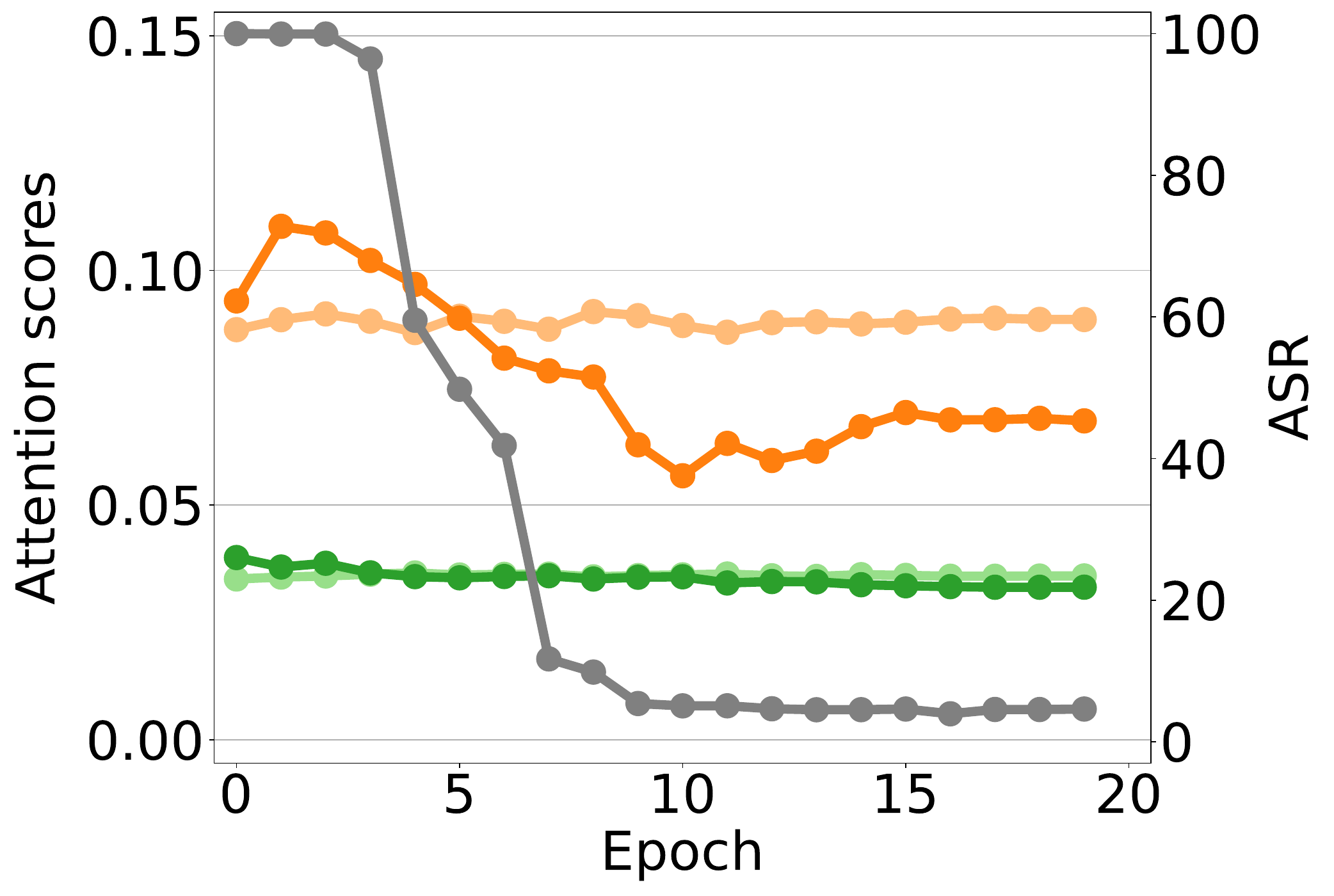}
        \vspace{-0.5cm}
        \caption{Prefix-tuning}
    \end{subfigure}
    \caption{PEFT training dynamics on SST-2 under the POR attack. The $L_2$-norms of the PEFT layers and those of the backdoored PLMs (\textit{left}). The average attention scores of trigger and normal tokens (\textit{right}).}
    \label{fig:training_dynamics}
    \vspace{-0.4cm}
\end{figure}

\section{Training Dynamics}
\label{appendi:training_dynamics}

To convince the effectiveness of our proposed techniques, we analyze the impact of neuron amplification and attention regularization during the training process, as illustrated in Figure~\ref{fig:training_dynamics}. 
We exclude the CACC for each result because its decrease is negligible (see Figure~\ref{fig:training}).


We evaluate the $L_2$-norms of the PEFT layers and the backdoored PLM layers (see Figure~\ref{fig:training_dynamics}~\textit{left}). Specifically, we present the norm of PEFT layers by comparing their values with or without our defense method.
Without any defense, the norm of the PEFT layers remain significantly lower than that of the PLM throughout training.
This is because the PEFT layers have been initialized with zero or minimal weights, which stabilizes training.
The observed decrease in ASR, corresponding with an increase in the norm of PEFT, implies that our defense method can neutralize backdoors that would have persisted due to low norms in the absence of a defense.
Despite increasing the norm of PEFT parameters, the models have been effectively trained on the downstream tasks.

In addition, we analyze the attention scores of trigger and normal tokens to the \texttt{[CLS]} token during training (see Figure~\ref{fig:training_dynamics}~\textit{right}). Without defense, the trigger tokens show abnormally higher attention scores compared to the normal ones throughout the training. By penalizing their influence, our defense method narrows the gap in attention scores, thereby effectively mitigating the backdoors.

\begin{table}[t]
    \centering
    \caption{Ablation study on RoBERTa the adapter models for SST-2, without the neuron amplification loss (w/o amp) and the attention regularization loss (w/o reg).}
    \footnotesize
        \begin{tabular} {@{ } c l @{\hspace{6pt}} c c c c @{ }}
            \toprule
            \textbf{Attack} & \textbf{Method} & \textbf{CACC} & \textbf{ASR} & \textbf{MASR} & \textbf{AASR} \\ \midrule
            \multirow{4}{*}{POR} & w/o amp & 91.21 & 12.34 & 11.61 & 5.95\\
                    & w/o reg & 92.53 & 5.34 & \textbf{2.91} & \textbf{2.14} \\ \cmidrule{2-6}
                    & \ours{} & 91.10 & \textbf{5.18} & 2.96 & 2.26 \\ \midrule
            \multirow{4}{*}{NeuBA}   & w/o amp & 93.47 & 40.48 & 65.20 & 19.14\\ 
                    & w/o reg & 93.08 & 10.09 & 9.64 & 4.47 \\ \cmidrule{2-6}
                    & \ours{} & 92.86 & \textbf{4.79} & \textbf{3.95} & \textbf{2.15} \\ \midrule
            \multirow{4}{*}{BadPre}   & w/o amp & 93.74 & 4.98 & 3.82 & 2.33 \\ 
                    & w/o reg & 93.57 & 13.38 & 19.56 & 9.78\\ \cmidrule{2-6}
                    & \ours{} & 93.96 & \textbf{2.75} & \textbf{1.73} & \textbf{1.49}  \\ \midrule
            \multirow{4}{*}{UOR}   & w/o amp & 90.17 & 22.53 & 40.29 & 8.84 \\ 
                    & w/o reg & 89.51 & 13.25 & 22.51 & 6.12 \\ \cmidrule{2-6}
                    & \ours{} & 89.51 & \textbf{6.38} & \textbf{8.08} & \textbf{2.65} \\
            \bottomrule
        \end{tabular}%
    \label{tab:ablation}
\end{table}

\begin{figure}[t]
    \centering
    \begin{subfigure}[t]{0.49\linewidth}
        \centering
        \includegraphics[width=\linewidth]{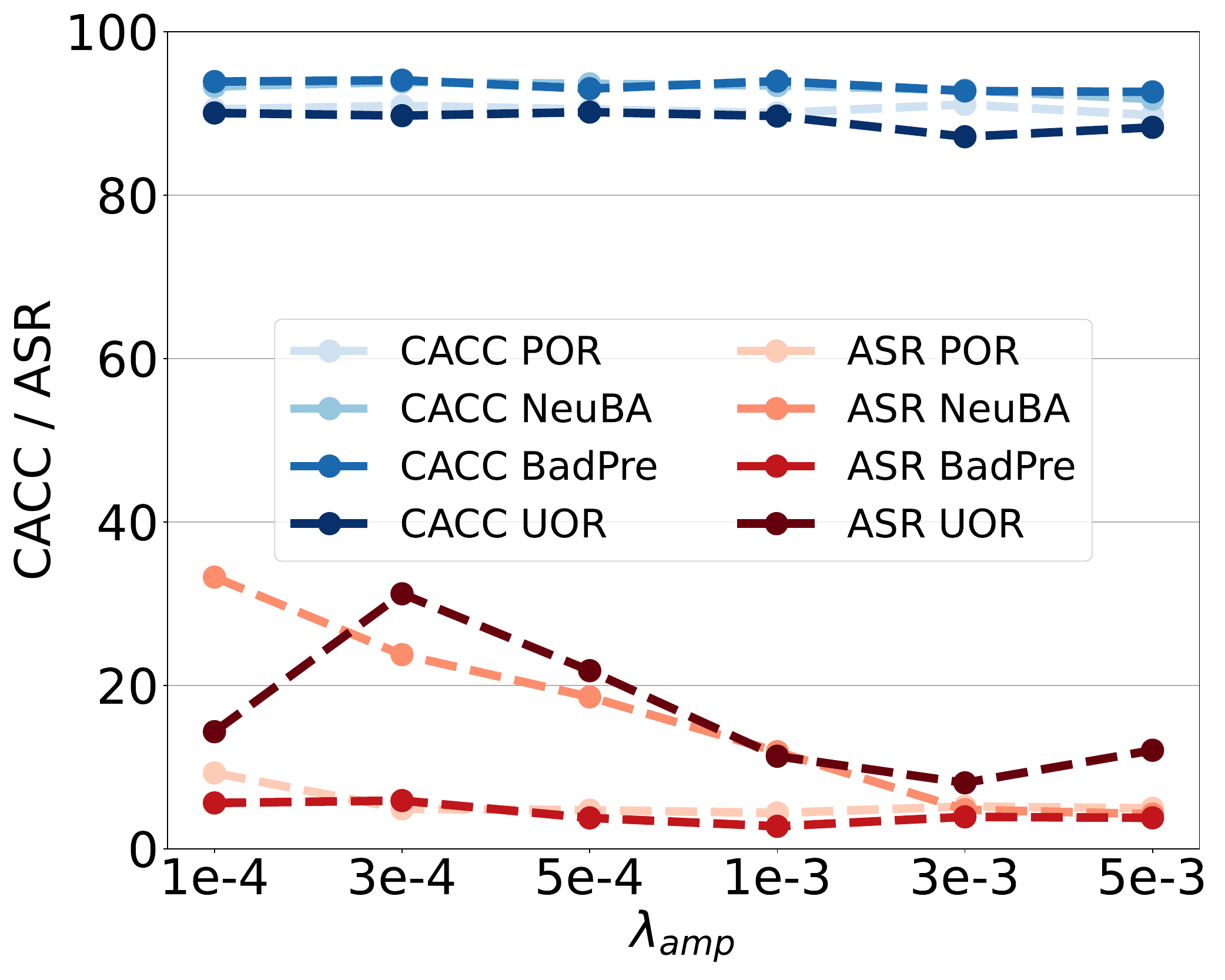}
        \caption{Coefficient of the neuron amplification loss.} 
        \vspace{2mm}
    \end{subfigure}
    \begin{subfigure}[t]{0.49\linewidth}
        \centering
        \includegraphics[width=\linewidth]{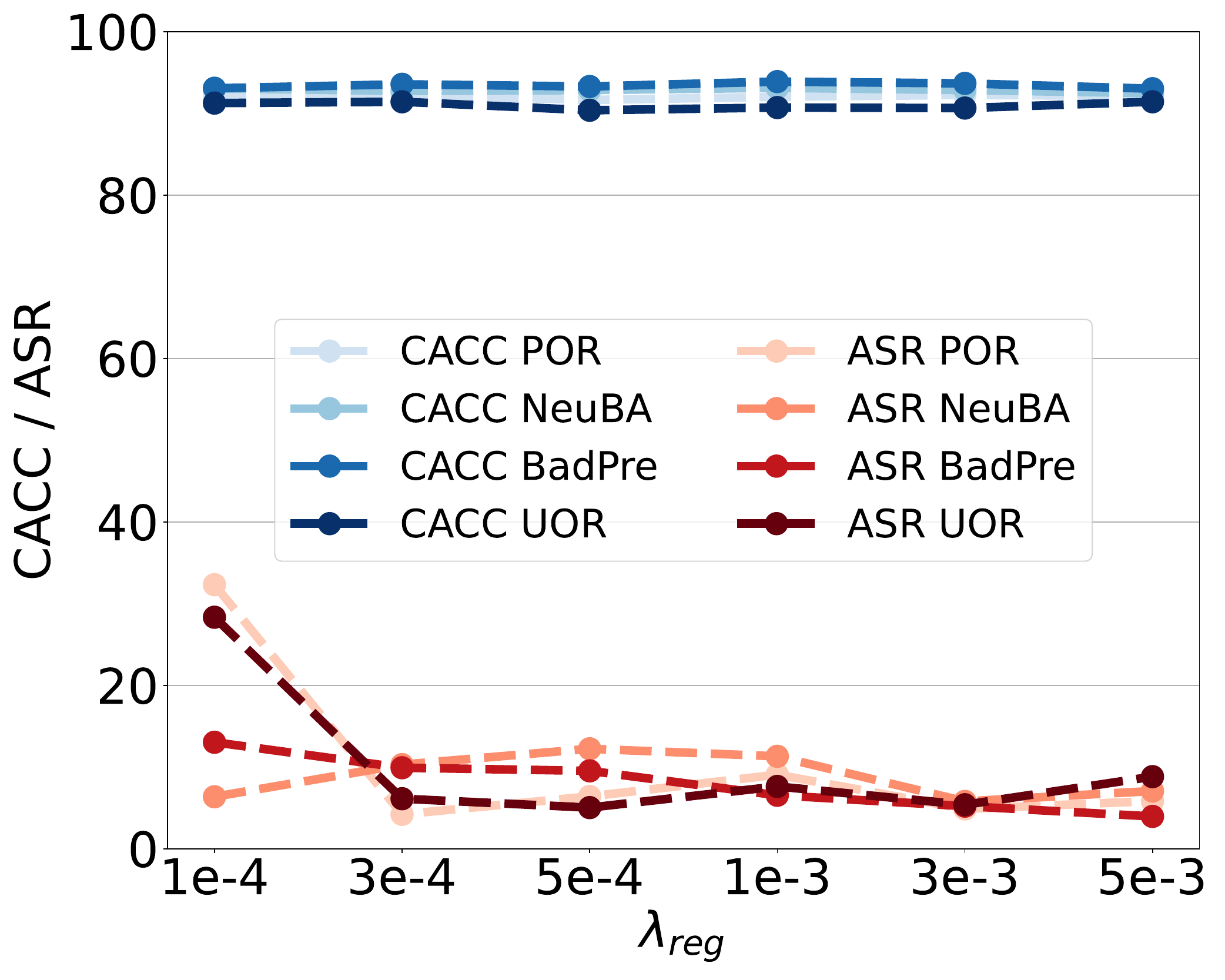}
        \caption{Coefficient of the attention regularization loss.}
    \end{subfigure}
    \caption{Defense performance of the adapter models using RoBERTa on SST-2 by adjusting defense loss coefficients.}
    \label{fig:coefficient}
\end{figure}

\section{Ablation Study}
\label{appendix:ablation_study}

We conduct an ablation study by removing the neuron amplification loss ($\mathcal{L}_{amp}$) or the attention regularization loss ($\mathcal{L}_{reg}$) from Equation~\ref{eq:loss}. The results are illustrated in Table~\ref{tab:ablation}.

Removing $\mathcal{L}_{amp}$ leads to a significant increase in ASR, indicating that amplifying the weights of matrices is crucial for eliminating backdoor information from their outputs. Particularly, $\mathcal{L}_{amp}$ plays a significant role in defending against attacks that target the \texttt{[CLS]} token, such as POR, NeuBA, and UOR. However, relying solely on $\mathcal{L}_{amp}$ for defense is not sufficient due to the limited number of parameters available for amplification.

On the other hand, the contribution of $\mathcal{L}_{reg}$ in neutralizing backdoors is also notable, except in the case of the POR attack. While it might penalize some non-trigger tokens, the minimal decrease in CACC when including $\mathcal{L}_{reg}$ suggests that such negative impacts are negligible. While the effectiveness of each loss varies depending on the attack, employing both $\mathcal{L}_{amp}$ and $\mathcal{L}_{reg}$ together offers the most comprehensive defense against a range of attacks.

\section{Impacts of Defense Loss Coefficients}
\label{appendix:impacts_of_defense_loss_coefficients}

To evaluate the effects of the neuron amplification and attention regularization losses, we analyze performance changes by adjusting $\lambda_{amp}$ and $\lambda_{reg}$. We present the results for adapter models using RoBERTa on the SST-2 dataset in Figure~\ref{fig:coefficient}. Adjusting $\lambda_{amp}$ reveals wide variations in ASR for NeuBA and UOR attacks, with the ASR generally decreasing as the coefficient is increased. Similarly, increasing $\lambda_{reg}$ results in a reduction in ASR. However, ASR values remain relatively unaffected by the coefficients. In both cases --- adjusting $\lambda_{amp}$ and $\lambda_{reg}$ --- the CACCs of backdoored models remain stable, even at high coefficient values, highlighting the reliability of our defense method.

\section{Defense Performance on Additional Classification Tasks}
\label{appendix:additional_classification_task}

We further evaluate our defense method on several classification tasks: natural language inference (NLI) -- SNLI~\cite{bowman2015large}, named entity recognition (NER) -- CoNLL 2003~\cite{sang2003introduction}, and question and answering (QA) -- SQuAD~\cite{rajpurkar2016squad}.

\noindent\textbf{Attack settings.} As the POR and NeuBA attacks target sentence classification tasks by manipulating the \texttt{[CLS]} output, we adapt these attacks to token classification tasks by forcing all the token outputs toward the adversarial representations. The method of the BadPre attack remains the same as that used for sentence classification tasks.

\noindent\textbf{Metrics.} For the NER task, we measure task performance on clean samples using the clean F1-score (\textbf{F1}). Additionally, we assess attack performance by the F1-score drop (\textbf{F1 drop}) when triggers are inserted. For the QA task, we evaluate performance using the clean exact match (\textbf{EM}) and clean F1-score (\textbf{F1}), along with the exact match drop (\textbf{EM drop}) and F1-score drop (\textbf{F1 drop}) to measure attack performance.

We present the defense performance for these three classification tasks in Table~\ref{tab:overall_roberta_plus}. For CoNLL 2003 and SQuAD, we ony compare results with ONION as RAP and PSIM are tailored to sentence classification tasks. According to the attack performance metrics, our defense method also demonstrates notable effectiveness in these advanced classification tasks. It shows exceptionally high defense performance in CoNLL 2003 with an average F1-drop of 6.01 and ASR of 1.21\%. This result aligns with the greater defense effectiveness observed in multiclass classification tasks in Section~\ref{sec:defense_performance}.

Similar to the observation in other sentence classification tasks, the defense performances of RAP and PSIM are unsatisfactory, except for the effectiveness of PSIM against the POR attack. In addition, ONION also struggles to provide effective defense for these advanced tasks; despite conservatively selected thresholds, it results in significant reductions in CACC and clean F1-score, particularly for SNLI and SQuAD. Our method, however, effectively defends with only minor degradation in clean F1-score, averaging 1.71 for CoNLL and 2.50 for SQuAD.

\section{Defense Performance: BERT}
\label{appendix:result_bert}

We present our experiments with BERT in Table~\ref{tab:overall_bert}. Consistent with the results from RoBERTa models, our defense method demonstrates significant effectiveness in protecting PEFT models against task-agnostic backdoors. On average, it achieves a 72.6\% reduction in ASR while only resulting in a slight decrease of 1.67\% in CACC. Compared to the baseline methods, ONION exhibits notable defense capabilities, particularly for the LoRA architectures in the SST-2 task. However, our method significantly outperforms both ONION and PSIM in almost all other cases.

\begin{table}[t]
    \centering
    \caption{Defense performance of the PEFT models using the large-version PLMs on SST-2 against the POR attack.}
    \footnotesize
        \begin{tabular} {@{} c @{ } l c @{\hspace{6pt}} c @{\hspace{6pt}} c @{\hspace{6pt}} c}
            \toprule
            \multirow{2}{*}{\begin{tabular}{c}\textbf{Model}\\\textbf{PEFT}\end{tabular}} & \multirow{2}{*}{\textbf{Method}} & \multirow{2}{*}{\textbf{CACC}} & \multirow{2}{*}{\textbf{ASR}} & \multirow{2}{*}{\textbf{MASR}} & \multirow{2}{*}{\textbf{AASR}} \\ 
            & & & \\\midrule
            \multirow{2}{*}{\begin{tabular}{c}RoBERTa\\Adapter\end{tabular}} & w/o def & 95.17 & 100 & 100 & 72.18\\
            & \ours{} & 94.56 & 2.21 & 1.66 & 1.20\\ \cmidrule{2-6}
            \multirow{2}{*}{\begin{tabular}{c}RoBERTa\\LoRA\end{tabular}} & w/o def & 95.06 & 98.04 & 100 & 72.20 \\
            & \ours{} & 93.68 & 3.99 & 2.73 & 1.86\\ \cmidrule{2-6}
            \multirow{2}{*}{\begin{tabular}{c}RoBERTa\\Prefix\end{tabular}} & w/o def & 93.35 & 100 & 100 & 99.35 \\
            & \ours{} & 93.03 & 2.95 & 1.81 & 1.42\\ \midrule
            \multirow{2}{*}{\begin{tabular}{c}BERT\\Adapter\end{tabular}} & w/o def & 92.48 & 100 & 100 & 71.87\\
            & \ours{} & 91.81 & 4.07 & 1.92 & 1.53\\ \cmidrule{2-6}
            \multirow{2}{*}{\begin{tabular}{c}BERT\\LoRA\end{tabular}} & w/o def & 93.30 & 100 & 100 & 98.32\\
            & \ours{} & 92.03 & 3.82 & 2.10 & 1.66 \\ \cmidrule{2-6}
            \multirow{2}{*}{\begin{tabular}{c}BERT\\Prefix\end{tabular}} & w/o def & 93.19 & 100 & 100 & 99.01 \\
            & \ours{} & 91.98 & 3.82 & 2.89 & 1.53\\
            \bottomrule
        \end{tabular}%
    \label{tab:large_model}
\end{table}

\section{Defense Performance of Large-version PLMs}
\label{appendix:result_large}

Both the base- and large- versions of the PLMs share the identical architecture (i.e., transformer layers). Additionally, all other training conditions, such as the training datasets, remain consistent across versions, although the optimal hyperparameters for defense may vary. Therefore, we expect our defense method to be still effective for the large-version PLMs. To support this, we conduct experiments on the large-version RoBERTa (\texttt{roberta-large}) and BERT (\texttt{bert-large-uncased}), with the results presented in Table~\ref{tab:large_model}. Our defense method significantly reduces the backdoor effect across all PEFT architectures for both PLMs.

\section{Defense against the UOR Attack}
\label{appendix:uor}

In Table~\ref{tab:overall_uor}, we present the performance evaluation of PEFT models using RoBERTa and BERT in defending against the UOR attack, an optimization-based task-agnostic backdoor. For models based on RoBERTa, we can successfully mitigate the backdoor attacks, performing better than the ONION and PSIM baselines. For BERT models, PSIM provides the most effective defense. However, our defense method also significantly lowers ASR in most cases. These results emphasize the practicality of our method in protecting against a range of attacks.

\begin{table*}[t]
    \centering
    \caption{Defence performance of the RoBERTa models on additional classification tasks.}
    \scriptsize
        \begin{tabular} {@{} c @{ } c c c c c c @{ } c c c c c c @{ } c c c c c @{\hspace{6pt}}}
            \toprule
            \multirow{3}{*}{\begin{tabular}{c} \textbf{Attack} \\ \textbf{PEFT} \end{tabular}} & \multirow{3}{*}{\textbf{Defense}} & \multicolumn{4}{c}{\textbf{SNLI}} & &  \multicolumn{5}{c}{\textbf{CoNLL 2003}} & & \multicolumn{4}{c}{\textbf{SQuAD}} \\  \cmidrule{3-6} \cmidrule{8-12} \cmidrule{14-17}
                              &  & CACC & ASR & MASR & AASR & & F1 & F1 drop & ASR & MASR & AASR & & EM &  F1 & EM drop & F1 drop \\  \midrule 
                \multirow{4}{*}{\begin{tabular}{c} POR \\ Adapter \end{tabular}} & w/o def & 81.10 & 100 & 100 & 86.79 & & 91.79 & 91.63 & 100 & 100 & 99.99 & & 73.25 & 83.25 & 66.50 & 66.32 \\
                             & ONION & 72.30 & 91.49 & 75.27 & 67.53 & & 89.02 & 7.76 & 22.34 & 17.28 & 10.93 & & 58.20 & 70.57 & 52.06 & 52.30 \\ 
                             & RAP & 78.85 & 100 & 100 & 77.42 & & - & - & - & - & - & & - & - & - & - \\ 
                             & PSIM & 81.55 & \textbf{0.00} & \textbf{0.00} & \textbf{0.00} & &  - & - & - & - & - & & - & - & - & - \\
                             \cmidrule{2-17}
                             & \ours{} & 80.60 & 6.76 & 3.28 & 2.11 & & 90.60 & \textbf{5.67} & \textbf{0.74} & \textbf{0.48} & \textbf{0.23} & & 72.65 & 82.56 & \textbf{8.92} & \textbf{7.29} \\ \cmidrule{2-17}
                \multirow{4}{*}{\begin{tabular}{c} POR \\ LoRA \end{tabular}} & w/o def & 79.75 & 100 & 100 & 98.06 & & 91.31 & 85.60 & 98.01 & 97.58 & 87.56 & & 75.50 & 84.91 & 63.76 & 58.83 \\
                             & ONION & 71.60 & 91.90 & 80.66 & 73.96 & & 88.77 & \textbf{5.62} & 17.52 & 17.82 & 12.96 & & 59.95 & 71.94 & 51.90 & 49.74\\ 
                             & RAP & 76.80 & 100 & 100 & 96.68 & & - & - & - & - & - & & - & - & - & - \\
                             & PSIM & 78.30 & \textbf{0.00} & \textbf{0.00} & \textbf{0.00} & &  - & - & - & - & - & & - & - & - & - \\ \cmidrule{2-17}
                             & \ours{} & 77.60 & 12.05 & 8.63 & 4.08 & & 89.83 & 6.20 & \textbf{0.69} & \textbf{0.39} & \textbf{0.22} & & 72.05 & 81.53 & \textbf{28.84} & \textbf{24.65}\\ \cmidrule{2-17}
                \multirow{4}{*}{\begin{tabular}{c} POR \\ Prefix \end{tabular}} & w/o def & 78.70 & 100 & 100 & 82.97 & & 91.40 & 87.89 & 100 & 100 & 96.82 & & 73.30 & 83.19 & 55.67 & 56.15  \\
                             & ONION & 71.15 & 93.46 & 78.69 & 64.87 & & 88.83 & 8.37 & 19.49 & 18.10 & 11.51 & & 60.90 & 72.20 & 46.01 & 46.38 \\ 
                             & RAP & 76.00 & 100 & 100 & 82.53 & & - & - & - & - & - & & - & - & - & - \\
                             & PSIM & 78.95 & 38.57 & 42.44 & 7.07 & &  - & - & - & - & - & & - & - & - & -  \\ \cmidrule{2-17}
                             & \ours{} & 78.85 & \textbf{5.90} & \textbf{2.11} & \textbf{1.51} & & 89.62 & \textbf{6.21} & \textbf{1.62} & \textbf{1.03} & \textbf{0.30} & & 71.70 & 82.05 & \textbf{27.46} & \textbf{26.14} \\ \midrule
                \multirow{4}{*}{\begin{tabular}{c} NeuBA \\ Adapter \end{tabular}}  & w/o def & 83.75 & 100 & 98.33 & 92.97 & & 91.57 & 86.89 & 100 & 100 & 92.18 & & 73.70 & 84.22 & 58.83 & 60.84\\
                             & ONION & 74.90 & 92.39 & 72.90 & 69.57 & & 88.77 & 5.61 & 16.32 & 18.06 & 13.51 & & 57.20 & 71.05 & 49.47 & 50.64 \\ 
                             & RAP & 81.45 & 99.88 & 94.46 & 80.75 & & - & - & - & - & - & & - & - & - & - \\
                             & PSIM & 84.95 & 100 & 98.04 & 92.73 & &  - & - & - & - & - & & - & - & - & -  \\ \cmidrule{2-17}
                             & \ours{} & 80.80 & \textbf{6.31} & \textbf{3.04} & \textbf{2.26} & & 90.05 & \textbf{5.12} & \textbf{0.73} & \textbf{0.44} & \textbf{0.23} & & 72.55 & 82.19 & \textbf{15.63} & \textbf{15.19}\\ \cmidrule{2-17}
                \multirow{4}{*}{\begin{tabular}{c} NeuBA \\ LoRA \end{tabular}}  & w/o def & 80.45 & 96.83 & 88.70 & 66.34 & & 90.96 & 80.10 & 91.01 & 99.40 & 76.09 & & 74.05 & 83.89 & 55.62 & 56.04\\
                             & ONION & 72.60 & 90.43 & 70.41 & 53.46 & & 88.43 & 5.83 & 16.10 & 18.15 & 13.42 & & 60.20 & 71.52 & 46.58 & 47.41\\
                             & RAP & 78.50 & 96.23 & 86.53 & 55.93 & & - & - & - & - & - & & - & - & - & - \\
                             & PSIM & 81.80 & 98.53 & 90.11 & 66.24 & &  - & - & - & - & - & & - & - & - & -  \\ \cmidrule{2-17}
                             & \ours{} & 79.10 & \textbf{8.41} & \textbf{4.27} & \textbf{2.61} & & 89.66 & \textbf{5.21} & \textbf{1.03} & \textbf{0.92} & \textbf{0.26} & & 70.00 & 80.76 & \textbf{23.62} & \textbf{22.23} \\ \cmidrule{2-17}
                \multirow{4}{*}{\begin{tabular}{c} NeuBA \\ Prefix \end{tabular}}  & w/o def & 84.60 & 100 & 94.41 & 89.89 & & 91.05 & 78.36 & 100 & 100 & 81.99 & & 74.40 & 83.87 & 47.65 & 47.52\\
                             & ONION & 74.85 & 91.52 & 70.36 & 64.73 & & 88.37 & \textbf{6.32} & 16.38 & 17.02 & 13.04 & & 61.85 & 71.82 & 41.35 & 41.73\\ 
                             & RAP & 81.85 & 100 & 93.09 & 85.15 & & - & - & - & - & - & & - & - & - & - \\
                             & PSIM & 84.75 & 100 & 94.28 & 89.80 & &  - & - & - & - & - & & - & - & - & -  \\ \cmidrule{2-17}
                             & \ours{} & 81.00 & \textbf{7.59} & \textbf{3.76} & \textbf{2.39} & & 87.42 & 8.49 & \textbf{4.02} & \textbf{3.30} & \textbf{0.73} & & 71.60 & 81.88 & \textbf{26.88} & \textbf{22.50}\\ \midrule
                \multirow{4}{*}{\begin{tabular}{c} BadPre \\ Adapter \end{tabular}}  & w/o def & 83.70 & 67.62 & 100 & 94.68 & & 91.39 & 85.31 & 90.98 & 95.64 & 92.07 & & 74.00 & 83.98 & 69.48 & 78.56 \\
                             & ONION & 74.25 & 63.64 & 73.86 & 69.70 & & 89.19 & 32.43 & 49.95 & 32.30 & 26.49 & & 60.50 & 72.22 & 51.61 & 56.08\\ 
                             & RAP & 81.75 & 67.56 & 99.31 & 93.75 & & - & - & - & - & - & & - & - & - & - \\
                             & PSIM & 84.45 & 65.72 & 100 & 94.02 & &  - & - & - & - & - & & - & - & - & -  \\ \cmidrule{2-17}
                             & \ours{} & 81.15 & \textbf{7.83} & \textbf{4.26} & \textbf{2.84} & & 89.96 & \textbf{5.75} & \textbf{0.67} & \textbf{0.21} & \textbf{0.12} & & 70.35 & 81.21 & \textbf{6.38} & \textbf{4.56} \\ \cmidrule{2-17}
                \multirow{4}{*}{\begin{tabular}{c} BadPre \\ LoRA \end{tabular}}  & w/o def & 83.35 & 64.01 & 100 & 93.11 & & 91.35 & 63.60 & 43.38 & 29.25 & 25.13 & & 73.65 & 83.57 & 54.74 & 58.15\\
                             & ONION & 74.75 & 69.63 & 73.41 & 67.08 & & 89.29 & \textbf{4.68} & 14.75 & 16.69 & 13.18 & & 58.70 & 70.56 & 46.20 & 48.36\\ 
                             & RAP & 81.15 & 63.38 & 99.91 & 93.02 & & - & - & - & - & - & & - & - & - & - \\
                             & PSIM & 85.45 & 66.18 & 100 & 93.56 & &  - & - & - & - & - & & - & - & - & -  \\ \cmidrule{2-17}
                             & \ours{} & 81.20 & \textbf{7.08} & \textbf{3.48} & \textbf{2.72} & & 89.61 & 5.93 & \textbf{0.77} & \textbf{0.42} & \textbf{0.22} & & 69.35 & 80.24 & \textbf{4.94} & \textbf{4.20} \\ \cmidrule{2-17}
                \multirow{4}{*}{\begin{tabular}{c} BadPre \\ Prefix \end{tabular}}  & w/o def & 84.45 & 64.59 & 100 & 93.59 & & 90.91 & 84.82 & 46.36 & 39.71 & 31.69 & & 75.05 & 84.30 & 41.92 & 37.49  \\
                             & ONION & 75.20 & 69.02 & 74.22 & 68.45 & & 88.69 & \textbf{4.39} & 14.37 & 18.01 & 14.00 & & 62.25 & 73.33 & 39.16 & 36.82\\
                             & RAP & 81.90 & 63.31 & 100 & 93.95 & & - & - & - & - & - & & - & - & - & - \\
                             & PSIM & 84.35 & 66.69 & 100 & 94.07 & & - & - & - & - & - & & - & - & - & -  \\ \cmidrule{2-17}
                             & \ours{} & 82.25 & \textbf{5.71} & \textbf{2.52} & \textbf{1.80} & & 89.53 & 5.51 & \textbf{0.58} & \textbf{0.23} & \textbf{0.12} & & 69.90 & 80.23 & \textbf{22.63} & \textbf{19.01} \\
            \bottomrule
        \end{tabular}%
    \label{tab:overall_roberta_plus}
\end{table*}

\begin{table*}[t]
    \centering
    \caption{Defense performance against backdoors in BERT models across PEFT architectures.}
    \scriptsize
        \begin{tabular} {@{} c @{ } c c c c c @{ } c c c c c @{ } c c c c c @{\hspace{6pt}}}
            \toprule
            \multirow{3}{*}{\begin{tabular}{c} \textbf{Attack} \\ \textbf{PEFT} \end{tabular}} & \multirow{3}{*}{\textbf{Defense}} & \multicolumn{4}{c}{\textbf{SST-2}} & &  \multicolumn{4}{c}{\textbf{AG News}} & & \multicolumn{4}{c}{\textbf{HSOL}} \\  \cmidrule{3-6} \cmidrule{8-11} \cmidrule{13-16}
                              &  & CACC & ASR & MASR & AASR & & CACC & ASR & MASR & AASR & & CACC & ASR & MASR & AASR \\  \midrule 
                \multirow{5}{*}{\begin{tabular}{c} POR \\ Adapter \end{tabular}} & w/o def & 90.33 & 100 & 100 & 92.89 & & 91.50 & 100 & 99.93 & 99.45 & & 91.40 & 100 & 100 & 99.70 \\
                             & ONION & 88.36 & 42.57 & 36.04 & 25.81 & & 90.00 & 15.72 & 6.59 & 4.96 & & 71.60 & 81.98 & 65.65 & 52.20 \\ 
                             & RAP & 86.93 & 69.12 & 73.71 & 49.72 & & 84.90 & 94.82 & 100 & 68.47 & & 89.55 & 99.78 & 89.79 & 74.02 \\
                             & PSIM & 90.28 & 48.45 & 100 & 33.12 & & 90.30 & 75.51 & 99.71 & 16.62 & & 91.05 & 99.95 & 100 & 66.33 \\  \cmidrule{2-16}
                             & \ours{}  & 89.18 & \textbf{4.00} & \textbf{2.43} & \textbf{1.82} & & 90.75 & \textbf{2.37} & \textbf{0.65} & \textbf{0.51} & & 91.30 & \textbf{3.07} & \textbf{5.41} & \textbf{3.82} \\ \cmidrule{2-16}
                \multirow{5}{*}{\begin{tabular}{c} POR \\ LoRA \end{tabular}} & w/o def & 90.94 & 100 & 100 & 99.98 & & 91.10 & 100 & 100 & 99.49 & & 91.55 & 100 & 100 & 99.83 \\
                             & ONION & 88.96 & \textbf{25.68} & \textbf{12.94} & \textbf{10.29} & & 89.30 & 15.06 & 5.52 & 4.77 & & 73.10 & 80.98 & 65.37 & 52.80 \\ 
                             & RAP & 86.05 & 94.08 & 97.39 & 61.01 & & 84.75 & 100 & 100 & 88.92 & & 89.10 & 99.95 & 98.90 & 85.50 \\
                             & PSIM & 90.01 & 99.94 & 100 & 33.31 & & 89.20 & 99.94 & 99.39 & 44.80 & & 91.00 & 100 & 100 & 66.62 \\  \cmidrule{2-16}
                             & \ours{} & 88.03 & 55.83 & 41.84 & 24.70 & & 89.40 & \textbf{7.38} & \textbf{2.57} & \textbf{1.12} & & 91.55 & \textbf{3.77} & \textbf{5.05} & \textbf{3.42} \\ \cmidrule{2-16}
                \multirow{5}{*}{\begin{tabular}{c} POR \\ Prefix \end{tabular}} & w/o def & 91.27 & 100 & 100 & 99.96 & & 91.30 & 100 & 99.93 & 93.84 & & 90.40 & 100 & 100 & 99.98 \\
                             & ONION & 89.35 & 58.39 & 45.34 & 37.36 & & 89.85 & 16.25 & 6.87 & 5.02 & & 70.00 & 80.86 & 66.22 & 52.42 \\ 
                             & RAP & 87.20 & 83.75 & 100 & 64.04 & & 85.85 & 100 & 100 & 94.51 & & 88.60 & 99.84 & 99.64 & 79.51 \\
                             & PSIM & 91.27 & 100 & 100 & 66.62 & & 90.10 & 100 & 99.86 & 73.11 & & 90.45 & 100 & 100 & 66.67 \\  \cmidrule{2-16}
                             & \ours{} & 89.02 & \textbf{17.46} & \textbf{27.35} & \textbf{7.01} & & 90.35 & \textbf{1.83} & \textbf{0.57} & \textbf{0.49} & & 91.70 & \textbf{1.47} & \textbf{3.59} & \textbf{1.85} \\ \midrule
                \multirow{5}{*}{\begin{tabular}{c} NeuBA \\ Adapter \end{tabular}}  & w/o def & 90.72 & 100 & 100 & 98.13 & & 91.75 & 96.95 & 94.24 & 49.84 & & 91.80 & 99.84 & 100 & 80.63 \\
                             & ONION & 88.85 & 19.53 & 9.10 & 6.73 & & 90.05 & 29.71 & 18.36 & 9.27 & & 72.45 & 80.06 & 63.45 & 45.54 \\ 
                             & RAP & 85.72 & 86.86 & 69.17 & 54.78 & & 85.60 & 69.24 & 99.67 & 31.98 & & 88.00 & 78.64 & 67.36 & 31.43 \\
                             & PSIM & 90.66 & 100 & 100 & 98.13 & & 90.20 & 97.13 & 95.06 & 50.15 & & 90.25 & 97.59 & 100 & 59.63 \\ \cmidrule{2-16}
                             & \ours{} & 88.14 & \textbf{10.09} & \textbf{5.70} & \textbf{4.04} & & 90.70 & \textbf{7.06} & \textbf{4.86} & \textbf{1.96} & & 91.00 & \textbf{3.35} & \textbf{4.64} & \textbf{2.26} \\ \cmidrule{2-16}
                \multirow{5}{*}{\begin{tabular}{c} NeuBA \\ LoRA \end{tabular}}  & w/o def & 90.12 & 100 & 100 & 99.07 & & 91.85 & 91.94 & 96.94 & 41.42 & & 91.55 & 91.37 & 85.48 & 57.99 \\
                             & ONION & 88.14 & \textbf{20.62} & \textbf{9.13} & \textbf{7.27} & & 90.05 & 10.66 & 4.39 & 2.31 & & 71.65 & 75.23 & 53.75 & 33.25 \\ 
                             & RAP & 85.78 & 97.93 & 100 & 76.51 & & 85.00 & 77.75 & 63.34 & 37.57 & & 88.10 & 74.24 & 75.38 & 24.18 \\
                             & PSIM & 88.36 & 100 & 100 & 65.76 & & 91.00 & 93.02 & 97.14 & 42.30 & & 90.30 & 90.63 & 84.74 & 57.19 \\ \cmidrule{2-16}
                             & \ours{} & 88.08 & 29.49 & 40.77 & 12.09 & & 89.60 & \textbf{5.92} & \textbf{2.65} & \textbf{1.22} & & 89.85 & \textbf{6.57} & \textbf{6.23} & \textbf{2.36} \\ \cmidrule{2-16}
                \multirow{5}{*}{\begin{tabular}{c} NeuBA \\ Prefix \end{tabular}}  & w/o def & 90.44 & 42.26 & 69.10 & 16.77 & & 90.65 & 69.66 & 78.19 & 26.55 & & 91.10 & 47.97 & 92.04 & 31.67 \\
                             & ONION & 88.63 & 20.38 & 23.09 & \textbf{7.26} & & 89.10 & 9.60 & 4.42 & 1.88 & & 71.10 & 42.05 & 37.14 & 18.37 \\ 
                             & RAP & 86.27 & 20.89 & 31.49 & 7.51 & & 85.75 & 35.44 & 81.17 & 24.24 & & 89.20 & 67.32 & 74.93 & 23.17 \\
                             & PSIM & 89.95 & 41.41 & 69.21 & 14.40 & & 89.45 & 71.30 & 78.13 & 22.70 & & 90.55 & 48.43 & 86.79 & 31.16 \\ \cmidrule{2-16}
                             & \ours{} & 88.36 & \textbf{17.78} & \textbf{21.52} & 8.94 & & 90.00 & \textbf{2.17} & \textbf{1.16} & \textbf{0.49} & & 92.00 & \textbf{0.76} & \textbf{1.65} & \textbf{1.27} \\ \midrule
                \multirow{5}{*}{\begin{tabular}{c} BadPre \\ Adapter \end{tabular}}  & w/o def & 91.54 & 50.15 & 100 & 100 & & 91.65 & 51.23 & 47.30 & 33.68 & & 92.55 & 81.85 & 92.35 & 60.07 \\
                             & ONION & 89.62 & 25.67 & 18.34 & 13.35 & & 89.95 & 9.01 & 3.17 & 2.14 & & 73.25 & 61.50 & 53.10 & 32.27 \\ 
                             & RAP & 85.94 & 39.05 & 48.33 & 45.85 & & 84.30 & 47.45 & 27.11 & 18.10 & & 89.35 & 29.26 & 17.01 & 9.94 \\
                             & PSIM & 89.95 & \textbf{0.00} & \textbf{0.00} & \textbf{0.00} & & 91.25 & 52.89 & 48.02 & 33.86 & & 91.30 & 80.91 & 92.58 & 59.53 \\ \cmidrule{2-16}
                             & \ours{} & 89.62 & 6.99 & 5.17 & 3.42 & & 90.65 & \textbf{3.53} & \textbf{2.15} & \textbf{1.10} & & 91.45 & \textbf{2.73} & \textbf{3.23} & \textbf{2.58} \\ \cmidrule{2-16}
                \multirow{5}{*}{\begin{tabular}{c} BadPre \\ LoRA \end{tabular}}  & w/o def & 90.39 & 51.64 & 100 & 99.92 & & 91.60 & 43.56 & 49.89 & 42.62 & & 91.20 & 84.21 & 77.67 & 54.90 \\
                             & ONION & 88.63 & 22.12 & \textbf{12.71} & 9.81 & & 90.35 & 8.69 & 2.39 & 1.87 & & 71.40 & 70.24 & 41.99 & 31.03 \\ 
                             & RAP & 86.99 & 41.19 & 57.41 & 48.27 & & 86.55 & 74.81 & 39.88 & 26.52 & & 89.00 & 23.76 & 18.40 & 15.58 \\
                             & PSIM & 89.57 & 51.64 & 100 & 99.92 & & 90.25 & 43.14 & 48.22 & 42.56 & & 90.50 & 83.46 & 75.29 & 54.15 \\ \cmidrule{2-16}
                             & \ours{} & 88.08 & \textbf{13.84} & 14.85 & \textbf{8.41} & & 89.35 & \textbf{3.36} & \textbf{1.09} & \textbf{0.84} & & 91.05 & \textbf{4.28} & \textbf{3.68} & \textbf{2.84} \\ \cmidrule{2-16}
                \multirow{5}{*}{\begin{tabular}{c} BadPre \\ Prefix \end{tabular}} & w/o def & 90.50 & 51.58 & 99.88 & 99.55 & & 91.65 & 56.36 & 70.44 & 60.60 & & 90.90 & 74.92 & 70.40 & 55.18 \\
                             & ONION & 88.85 & 34.86 & 33.49 & 25.37 & & 89.90 & 10.46 & 3.73 & 2.59 & & 71.25 & 58.60 & 49.43 & 35.47 \\ 
                             & RAP & 87.81 & 50.36 & 80.26 & 72.72 & & 85.75 & 66.85 & 62.34 & 44.19 & & 89.40 & 51.60 & 58.86 & 46.81 \\
                             & PSIM & 90.39 & 51.58 & 99.88 & 99.55 & & 90.75 & 56.55 & 71.05 & 60.35 & & 90.50 & 74.07 & 71.06 & 56.92 \\ \cmidrule{2-16}
                             & \ours{} & 89.24 & \textbf{5.78} & \textbf{5.02} & \textbf{2.57} & & 90.15 & \textbf{1.44} & \textbf{0.52} & \textbf{0.36} & & 91.65 & \textbf{2.07} & \textbf{4.29} & \textbf{3.30} \\
            \bottomrule
        \end{tabular}%
    \label{tab:overall_bert}
\end{table*}

\begin{table*}[t]
    \centering
    \caption{Defense performance against the UOR attack.}
    \scriptsize
        \begin{tabular} {@{} c @{ } c @{ } c c c c c @{ } c c c c c @{ } c c c c c @{\hspace{6pt}}}
            \toprule
            \multirow{3}{*}{\textbf{Model}} & \multirow{3}{*}{\begin{tabular}{c} \textbf{Attack} \\ \textbf{PEFT} \end{tabular}} & \multirow{3}{*}{\textbf{Defense}} & \multicolumn{4}{c}{\textbf{SST-2}} & &  \multicolumn{4}{c}{\textbf{AG News}} & & \multicolumn{4}{c}{\textbf{HSOL}} \\  \cmidrule{4-7} \cmidrule{9-12} \cmidrule{14-17}
                              & &  & CACC & ASR & MASR & AASR & & CACC & ASR & MASR & AASR & & CACC & ASR & MASR & AASR \\  \midrule 
                \multirow{15}{*}{RoBERTa} & \multirow{5}{*}{\begin{tabular}{c} UOR \\ Adapter \end{tabular}}  & w/o def & 91.82 & 53.59 & 85.66 & 37.46 & & 90.85 & 99.83 & 99.78 & 71.70 & & 90.70 & 99.78 & 100 & 80.82 \\
                             & & ONION & 89.95 & 30.46 & 25.12 & 14.48 & & 89.05 & 13.64 & 5.89 & 3.49 & & 78.30 & 71.33 & 61.54 & 40.47 \\ 
                             & & RAP & 88.19 & 32.89 & 30.30 & 17.87 & & 83.20 & 92.49 & 100 & 65.97 & & 88.25 & 97.07 & 87.72 & 45.82 \\
                             & & PSIM  & 91.60 & 53.47 & 85.66 & 30.25 & & 89.55 & 75.95 & 97.49 & 35.73 & & 89.70 & 99.67 & 100 & 48.57 \\ \cmidrule{3-17}
                             & & \ours{} & 89.51 & \textbf{6.38} & \textbf{8.08} & \textbf{2.65} & & 90.85 & \textbf{3.19} & \textbf{1.71} & \textbf{0.72} & & 91.80 & \textbf{2.29} & \textbf{3.15} & \textbf{2.37} \\ \cmidrule{3-17}
                & \multirow{5}{*}{\begin{tabular}{c} UOR \\ LoRA \end{tabular}}  & w/o def & 90.12 & 12.92 & 13.44 & 6.27 & & 90.70 & 96.69 & 84.57 & 42.73 & & 89.55 & 28.36 & 98.00 & 39.35 \\
                             & & ONION & 88.36 & 11.44 & \textbf{5.13} & 4.16 & & 88.85 & 11.71 & \textbf{3.55} & 2.50 & & 71.85 & 37.44 & 55.78 & 23.04 \\ 
                             & & RAP & 87.31 & 6.76 & 6.78 & 3.12 & & 86.40 & 23.93 & 97.40 & 33.94 & & 88.45 & 71.84 & 57.60 & 23.26 \\
                             & & PSIM & 89.13 & \textbf{6.34} & 5.92 & \textbf{2.53} & & 89.45 & 80.82 & 83.76 & 30.98 & & 90.20 & 30.54 & 98.48 & 39.88 \\ \cmidrule{3-17}
                             & & \ours{} & 90.72 & 8.84 & 8.59 & 4.12 & & 91.50 & \textbf{5.57} & 4.21 & \textbf{1.24} & & 91.50 & \textbf{6.01} & \textbf{7.52} & \textbf{3.63} \\ \cmidrule{3-17}
                & \multirow{5}{*}{\begin{tabular}{c} UOR \\ Prefix \end{tabular}}  & w/o def & 89.84 & 79.83 & 100 & 36.36 & & 91.55 & 99.62 & 99.36 & 57.91 & & 91.90 & 99.67 & 100 & 77.10 \\
                             & & ONION & 88.08 & 16.96 & 9.15 & 6.61 & & 89.70 & 12.32 & \textbf{5.26} & 3.30 & & 70.65 & 80.47 & 61.43 & 40.75 \\ 
                             & & RAP & 86.93 & 78.48 & 98.82 & 35.32 & & 85.45 & 97.65 & 100 & 70.33 & & 87.25 & 90.84 & 84.39 & 50.15 \\
                             & & PSIM & 89.62 & 79.83 & 100 & 36.36 & & 89.75 & 61.83 & 80.07 & 24.49 & & 90.20 & 99.29 & 100 & 47.32 \\ \cmidrule{3-17}
                             & & \ours{} & 88.47 & \textbf{5.83} & \textbf{3.85} & \textbf{2.73} & & 89.55 & \textbf{8.65} & 10.29 & \textbf{1.95} & & 90.50 & \textbf{7.29} & \textbf{32.35} & \textbf{11.89} \\  \midrule
                \multirow{15}{*}{BERT} & \multirow{5}{*}{\begin{tabular}{c} UOR \\ Adapter \end{tabular}} & w/o def & 90.17 & 94.64 & 100 & 61.47 & & 90.70 & 100 & 100 & 88.02 & & 91.25 & 100 & 100 & 76.72 \\
                             & & ONION & 88.30 & 21.21 & 12.31 & 7.44 & & 89.85 & 15.41 & 6.50 & 4.59 & & 79.05 & 70.52 & 63.57 & 36.45 \\ 
                             & & RAP & 86.60 & 68.03 & 59.14 & 32.39 & & 83.55 & 98.17 & 97.26 & 55.56 & & 89.45 & 99.73 & 94.99 & 62.88 \\
                             & & PSIM & 89.13 & \textbf{0.00} & \textbf{0.00} & \textbf{0.00} & & 88.80 & \textbf{0.00} & \textbf{0.00} & \textbf{0.00} & & 89.25 & \textbf{0.00} & \textbf{0.00} & \textbf{0.00} \\ \cmidrule{3-17}
                             & & \ours{} & 88.74 & 9.59 & 9.69 & 4.80 & & 90.15 & 6.27 & 5.73 & 1.56 & & 90.65 & 18.26 & 82.41 & 15.16 \\ \cmidrule{3-17}
                & \multirow{5}{*}{\begin{tabular}{c} UOR \\ LoRA \end{tabular}} & w/o def & 91.32 & 68.91 & 73.00 & 42.99 & & 91.20 & 87.50 & 99.49 & 43.21 & & 90.85 & 100 & 100 & 70.25 \\
                             & & ONION & 89.51 & \textbf{29.08} & \textbf{21.45} & 13.13 & & 89.90 & 11.18 & 4.89 & 2.52 & & 77.85 & 72.90 & 54.88 & 31.52 \\ 
                             & & RAP & 85.34 & 30.19 & 29.46 & 13.80 & & 85.00 & 93.68 & 64.49 & 39.48 & & 88.40 & 72.47 & 70.32 & 26.80 \\
                             & & PSIM & 89.73 & 33.13 & 67.69 & \textbf{11.28} & & 89.55 & \textbf{0.00} & \textbf{0.00} & \textbf{0.00} & & 89.15 & \textbf{0.00} & \textbf{0.00} & \textbf{0.00} \\ \cmidrule{3-17}
                             & & \ours{} & 88.63 & 33.02 & 34.40 & 14.87 & & 89.20 & 5.21 & 1.60 & 1.22 & & 91.30 & 6.90 & 9.73 & 3.80 \\ \cmidrule{3-17}
                & \multirow{5}{*}{\begin{tabular}{c} UOR \\ Prefix \end{tabular}} & w/o def & 90.55 & 69.19 & 99.25 & 34.34 & & 90.55 & 99.89 & 100 & 80.98 & & 91.90 & 100 & 100 & 80.07 \\
                             & & ONION & 88.85 & 34.49 & 39.00 & 13.67 & & 89.50 & 15.81 & 6.34 & 4.41 & & 80.15 & 66.13 & 65.02 & 47.22 \\ 
                             & & RAP & 85.89 & 50.58 & 75.72 & 22.05 & & 87.20 & 100 & 100 & 74.76 & & 88.60 & 99.89 & 99.93 & 75.98 \\
                             & & PSIM & 90.39 & \textbf{3.09} & \textbf{6.35} & \textbf{1.06} & & 89.30 & \textbf{0.00} & \textbf{0.00} & \textbf{0.00} & & 91.35 & \textbf{0.00} & \textbf{0.00} & \textbf{0.00} \\ \cmidrule{3-17}
                             & & \ours{} & 88.69 & 49.78 & 91.65 & 21.16 & & 90.40 & 1.83 & 0.80 & 0.48 & & 91.55 & 15.95 & 71.86 & 13.87 \\
            \bottomrule
        \end{tabular}%
    \label{tab:overall_uor}
\end{table*}


\end{document}